\RequirePackage[2020-02-02]{latexrelease}
\documentclass{article} 
\usepackage{iclr2022_conference,times}
\usepackage[utf8]{inputenc} 
\usepackage[T1]{fontenc}    
\usepackage{hyperref}       
\usepackage{url}            
\usepackage{booktabs}       
\usepackage{nicefrac}       
\usepackage{microtype}      
\usepackage{xcolor}         
\usepackage{amssymb}
\usepackage{amsmath}
\usepackage{amsfonts}       
\usepackage{graphicx,epstopdf}
\usepackage{tabularx}
\usepackage{subfigure}
\usepackage{enumitem}	
\usepackage{listings}
\usepackage{changepage}
\usepackage{pifont}
\usepackage{multirow}
\usepackage{wrapfig}
\usepackage{color,soul}
\usepackage{tikz}
\usepackage[normalem]{ulem}
\usepackage{pifont}
\usepackage{soul}
\usepackage{xspace}
\usepackage{graphicx}
\usepackage{multirow}
\usepackage{multicol}
\usepackage[printwatermark]{xwatermark}
\usepackage{pifont}
\usepackage{epsfig}
\usepackage{tikz}
\usepackage{filecontents}
\usepackage[toc,titletoc,page]{appendix}
\usepackage{graphicx}
\usepackage{varwidth}
\usepackage{capt-of}
\usepackage{comment}
\usepackage{floatrow}
\usepackage{bbm}

\newcommand*\circleb[1]{\tikz[baseline=(char.base)]{
    \node[shape=circle,fill=black,draw,text=white,inner sep=1pt,minimum height=4pt] (char) {\vphantom{WAH1g}#1};}}

\newcommand{\Paragraph}[1]{\vskip 3pt\noindent\textbf{#1 }}	

\makeatletter \DeclareRobustCommand\onedot{\futurelet\@let@token\@onedot}
\def\@onedot{\ifx\@let@token.\else.\null\fi\xspace}

\def\eg{\emph{e.g}\onedot} 
\def\ie{\emph{i.e}\onedot}

\def\etal{\emph{et al}\onedot}
\makeatother

\newenvironment{myitemize}%
  {\begin{itemize}
	[leftmargin=0cm,
		itemindent=.3cm,
		labelwidth=\itemindent,
		labelsep=0pt,
		parsep=3pt,
		topsep=2pt,
		itemsep=1pt,
		align=left]
  }%
  {\end{itemize}}

  {\begin{enumerate}
	[leftmargin=0cm,itemindent=.5cm,labelwidth=\itemindent,
		labelsep=0pt,
		parsep=1pt,
		topsep=1pt,
		itemsep=3pt,
		align=left]
  }%
  {\end{enumerate}}

\newcommand{\ours}{CarM\xspace}
\newcommand{\oursfull}{Carousel Memory\xspace}

\usepackage{arydshln}
\makeatletter
\def\adl@drawiv#1#2#3{%
        \hskip.5\tabcolsep
        \xleaders#3{#2.5\@tempdimb #1{1}#2.5\@tempdimb}%
                #2\z@ plus1fil minus1fil\relax
        \hskip.5\tabcolsep}
\newcommand{\cdashlinelr}[1]{%
  \noalign{\vskip\aboverulesep
           \global\let\@dashdrawstore\adl@draw
           \global\let\adl@draw\adl@drawiv}
  \cdashline{#1}
  \noalign{\global\let\adl@draw\@dashdrawstore
           \vskip\belowrulesep}}
\makeatother

\usepackage{amsmath,amsfonts,bm}

\def\eqref#1{equation~\ref{#1}}

\def\1{\bm{1}}

\DeclareMathAlphabet{\mathsfit}{\encodingdefault}{\sfdefault}{m}{sl}
\SetMathAlphabet{\mathsfit}{bold}{\encodingdefault}{\sfdefault}{bx}{n}

\usepackage{hyperref}
\usepackage{url}

\title{\oursfull: Rethinking the Design of Episodic Memory for Continual Learning}

\author{Antiquus S.~Hippocampus, Natalia Cerebro \& Amelie P. Amygdale \thanks{ Use footnote for providing further information
about author (webpage, alternative address)---\emph{not} for acknowledging
funding agencies.  Funding acknowledgements go at the end of the paper.} \\
Department of Computer Science\\
Cranberry-Lemon University\\
Pittsburgh, PA 15213, USA \\
\texttt{\{hippo,brain,jen\}@cs.cranberry-lemon.edu} \\
\And
Ji Q. Ren \& Yevgeny LeNet \\
Department of Computational Neuroscience \\
University of the Witwatersrand \\
Joburg, South Africa \\
\texttt{\{robot,net\}@wits.ac.za} \\
\AND
Coauthor \\
Affiliation \\
Address \\
\texttt{email}
}

\begin{document}

\maketitle
\renewcommand{\thefootnote}{\fnsymbol{footnote}}
\footnotetext[1]{Corresponding author (E-mail : \href{mjjeon@unist.ac.kr}{mjjeon@unist.ac.kr})}

\begin{abstract}
Continual Learning (CL) is an emerging machine learning paradigm that aims to learn from a continuous stream of tasks without forgetting knowledge learned from the previous tasks. 
To avoid performance decrease caused by forgetting, prior studies exploit episodic memory (EM), which stores a subset of the past observed samples while learning from new non-\emph{i.i.d.} data. 
Despite the promising results, since CL is often assumed to execute on mobile or IoT devices, the EM size is bounded by the small hardware memory capacity and makes it infeasible to meet the accuracy requirements for real-world applications.
Specifically, all prior CL methods discard samples overflowed from the EM and can never retrieve them back for subsequent training steps, incurring loss of information that would exacerbate catastrophic forgetting.
We explore a novel hierarchical EM management strategy to address the forgetting issue.
In particular, in mobile and IoT devices, real-time data can be stored not just in high-speed RAMs but in internal storage devices as well, which offer significantly larger capacity than the RAMs. 
Based on this insight, we propose to exploit the abundant storage to preserve past experiences and alleviate the 
forgetting by allowing CL to efficiently migrate samples between memory and storage without being interfered by the slow access speed of the storage.
We call it {\bf \oursfull (\ours)}.
As \ours is complementary to existing CL methods, we conduct extensive evaluations of our method with seven popular CL methods and show that \ours significantly improves the accuracy of the methods across different settings by large margins
in final average accuracy (up to 28.4\%) while retaining the same training efficiency.
\end{abstract}

\section{Introduction}
\label{sec:Introduction}

With the rising demand for realistic on-device machine learning, recent years have
witnessed a novel learning paradigm, namely continual learning (CL), for
training neural networks (NN) with a stream of non-\emph{i.i.d.}
data. In such a paradigm, the neural network is incrementally learned
with insertions of new tasks (\eg, a set of
classes)~\citep{icarl}. The NN model is expected to
continuously learn new knowledge from new tasks over time while retaining previously learned knowledge, which is a closer representation of how intelligent systems operate in the real world.
In this learning setup, the knowledge should be acquired not only from the
new data timely but also in a computationally efficient manner.
In this regard, CL is suitable for learning on mobile and IoT devices~\citep{Hayes2020REMINDYN, e2train}.

However, CL faces significant challenges from the notorious catastrophic
forgetting problem---knowledge learned in the past fading away as the NN model continues to learn new tasks~\citep{mccloskeyC89}.
Among many prior approaches to addressing this issue,
episodic memory (EM) is one of
the most effective approaches~\citep{darker, AGEM, tiny, LopezPaz2017GradientEM, gdumb}. EM is an \emph{in-memory} buffer that
stores old samples and replays them periodically while training 
models with new samples. EM needs to have a sufficiently large capacity to achieve a desired accuracy, and such capacity in need may vary significantly since incoming data may contain a varying number of tasks and classes at
different time slots and geo-locations~\citep{rainbow}. However, in practice, the size of EM is often quite small, bounded by \emph{limited} on-device memory capacity. The limited EM size makes it difficult to store a large number of samples or scale up to a
large number of tasks, preventing CL models from achieving high accuracy as training moves forward. To address the forgetting problem, we introduce a hierarchical EM method, which significantly enhances the effectiveness of episodic memory. Our method is motivated by the fact that  
modern mobile and IoT devices are commonly equipped with a deep memory
hierarchy including small memory with fast access (50--150~ns) and large
storage with slow access (25--250~$\mu$s), which is typically orders of
magnitude larger than the memory.
Provided by these different hardware characteristics, the memory is an ideal place to access samples at high speed
during training, promising short training time.
In contrast, the storage is an ideal place to store a significantly large number of old samples and use them for greatly improving model accuracy. 
The design goal of our scheme, \textbf{\oursfull} or \textbf{\ours}, is to combine the best of both
worlds to improve the episodic memory capacity by leveraging on-device storage but without significantly prolonging traditional memory-based CL approaches.

\ours stores as many observed samples as possible so long as it does not exceed a given storage capacity
(rather than discarding those overflowed from EM as done in existing
methods) and updates the in-memory EM while the model is still
learning with samples already in EM. 
One key research question is how to manage samples across EM and storage for both system efficiency and model accuracy. 
Here we propose a hierarchical memory-aware \emph{data swapping}, an \emph{online}
process that dynamically replaces a subset of in-memory samples used for model training with other
samples stored in storage, with an optimization goal in two
aspects. \underline{(1) System efficiency}. Prior single-level memory-only training approaches promise
timely model updates even in the face of real-time data that arrives with
high throughput. 
Therefore, we expect drawing old samples from slow storage
does not incur significant I/O overhead that affects the overall system efficiency, especially for mobile
and IoT devices. 
\underline{(2) Model accuracy}. 
\ours significantly increases the effective EM size, mitigating forgetting issues by avoiding important information from overflowing due to limited memory capacity. 
As a result, we expect our approach also improves the model accuracy by exploiting data samples preserved in the storage more effectively for training. To fulfill the competing goals, we design \ours from two different perspectives:
\textbf{swapping mechanism} (Section~\ref{sec:SystemDesign}) and \textbf{swapping policy} (Section~\ref{sec:Algorithms}). The swapping mechanism of \ours ensures that the slow speed of accessing the storage
does not become a bottleneck of continual model training by carefully hiding sample swapping latency through \emph{asynchrony}. 
Moreover, we propose various swapping policies to decide which and when to swap samples and incorporate them into a single component, namely, \emph{gate function}.
The gate function allows for fewer swapping samples, making \ours to march with low I/O bandwidth storage which is common for mobile and IoT devices.

One major benefit of \ours is that it is largely complementary to existing
episodic memory-based CL methods. By exploiting the memory hierarchy, we show that
\ours helps boost the accuracy of many existing methods by up to 28.4\% for
DER++~\citep{darker} in Tiny-ImageNet
dataset (Section~\ref{sec:Results}) and even allows them to retain their
accuracy with much smaller EM sizes.

With \ours as a strong baseline for episodic memory-based CL methods, some
well-known algorithmic optimizations may need to be revisited to ensure that
they are not actually at odds with data swapping. For example, we observe
that iCaRL~\citep{icarl}, BiC~\citep{bic}, and DER++~\citep{darker}, which
strongly depend on knowledge distillation for old tasks, can deliver higher
accuracy with \ours by limiting the weight coefficient on the distillation loss as a
small value in calculating training loss.
With \ours, such weight coefficient does not indeed necessarily be high or managed complicatedly as done in prior work, because we could now leverage a large amount of data in storage (with ground truth) to facilitate training performance.

\section{Related Work}
\label{sec:Related}

\Paragraph{Class incremental learning.}
The performance of CL algorithms heavily depends on scenarios and setups, as summarized by Van de Ven~\etal~\citep{ven2018three}.
Among them, we are particularly interested in class-incremental learning (CIL), where
task IDs are not given during inference~\citep{gepperthH16}.
Many prior proposals are broadly divided into two categories, rehearsal-based and regularization-based.
In rehearsal-based approaches, episodic memory stores a few samples of old tasks to replay in the future~\citep{rainbow,castro2018eccv,rwalk,icarl}.
On the contrary, regularization-based approaches exploit the information of old tasks implicitly retained in the model parameters, without storing samples representing old tasks~\citep{Kirkpatrick2017OvercomingCF,Zenke2017ContinualLT,Liu2018RotateYN,Li2017LearningWF,Lee2017OvercomingCF,Mallya2018PiggybackAA}.
As rehearsal-based approaches generally have shown the better performance in CIL~\citep{gdumb}, 
we aim to alleviate current drawbacks of the approaches (\ie, limited memory space) by incorporating data management across the memory-storage hierarchy.

The CIL setup usually assumes that the tasks contain disjoint set of classes~\citep{icarl,castro2018eccv,gepperthH16}. 
More recent studies introduce methods to learn from the blurry stream of tasks, where some samples across the tasks overlap in terms of class ID~\citep{gss,gdumb}.
Moreover, prior works can be classified as either \textit{offline}~\citep{bic, icarl, rwalk, castro2018eccv}, which allows a buffer to store incoming samples for the current task, or \textit{online}~\citep{Fini2020OnlineCL, gss, gmed}, which has no such buffer: a few works consider both~\citep{gdumb}.
Both online and offline methods can take advantage of \ours as our work focuses on improving episodic memory with a storage device.

\Paragraph{Episodic memory management.}
There are numerous episodic memory management strategies proposed in the literature~\citep{parisiKPK18} such as herding selection~\citep{Welling2009herding}, discriminative sampling~\citep{mnemonics}, entropy-based sampling~\citep{rwalk} and diversity-based sampling~\citep{kangJNC2020confcal,rainbow}.
A number of works have been proposed to compose the episodic memory with representative and discriminative samples.
Liu~\etal propose a strategy to store samples representing the mean and boundary of each class distribution~\citep{mnemonics}.
Borsos~\etal propose a coreset generation method using cardinality-constrained bi-level optimization~\citep{Borsos2020CoresetsVB}.
Cong~\etal propose a GAN-based memory aiming to perturb styles of remembered samples for incremental learning~\citep{Cong2020GANMW}.
Bang~\etal propose a strategy to promote the diversity of samples in the episodic memory~\citep{rainbow}.
These recent works improve the quality of the samples stored in the memory at the expense of excessive computation or difficulty~\citep{Borsos2020CoresetsVB} involved in training a generation model for perturbation~\citep{Cong2020GANMW,Borsos2020CoresetsVB}.
Interestingly, most of strategies show marginal accuracy improvements over the uniform random sampling despite the computational complexity~\citep{rwalk,castro2018eccv,icarl}. 
Other than sampling, there are works to generate samples of past tasks~\citep{Shin2017ContinualLW,Seff2017ContinualLI,Wu2018MemoryRG,hu2018overcoming}.
Unlike these works addressing the sampling efficiency, we focus on the systematically efficient method to manage
samples across the system memory hierarchy.

\Paragraph{Memory over-commitment in NN training.}
Prior work studies using storage or slow memory (\eg, host memory) as an extension of fast memory (\eg, GPU memory) to increase memory capacity for NN training~\citep{vDNN_micro16,SuperNeurons_ppopp18,AutoTM_asplos20,SwapAdvisor_asplos20,Capuchin_asplos20,Layrub,sentinel}. 
However, most of these works target at optimizing the conventional offline learning scenarios by swapping optimizer states, activations, or model weights between the fast memory and slow memory (or storage), whereas we focus on swapping samples in between episodic memory and storage to tackle the forgetting problem in the context of continual learning.  
In more general context, memory-storage caching has been studied to reduce memory and energy consumption for various applications~\citep{pacman,cache-less,spark}, which is orthogonal to our work.

\section{Proposed Method: \oursfull}
\label{sec:Approach}

\begin{figure}[!t]
\center
\includegraphics[width=\linewidth]{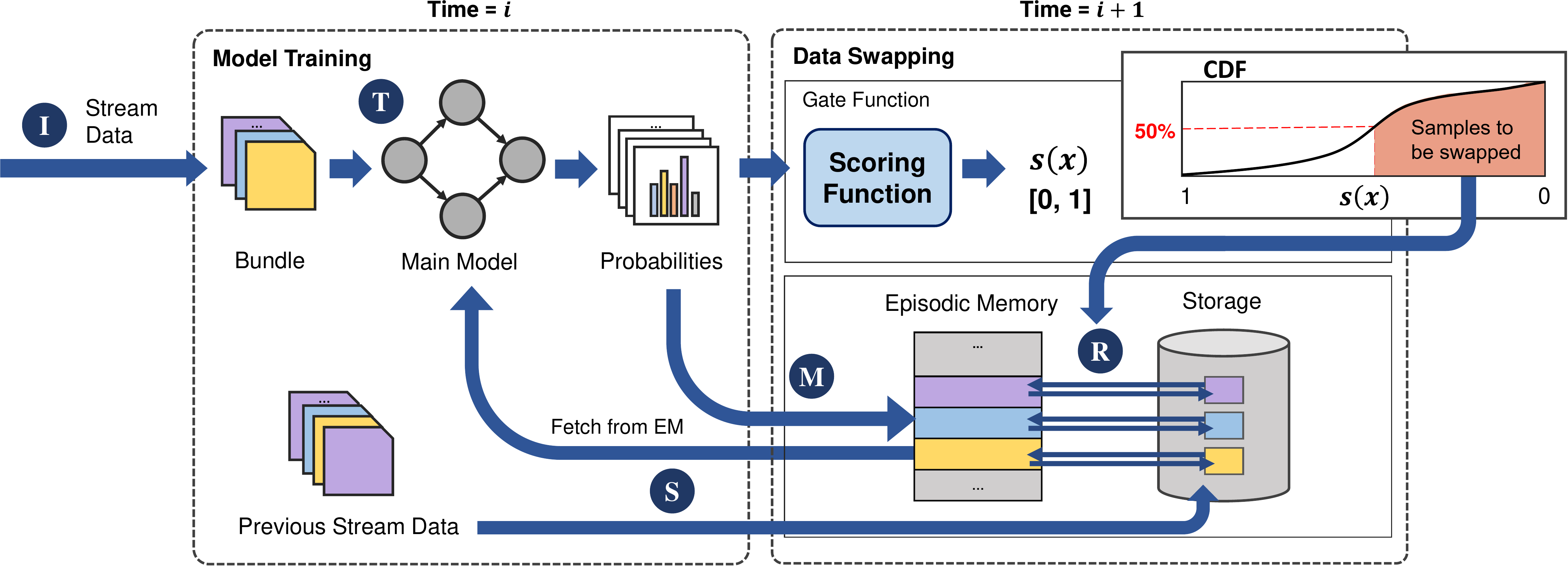}
\caption{Architecture and execution stages of the proposed \oursfull.
Swap worker replaces in-memory samples used for training with other samples stored in storage by leveraging gate function. }
\vspace{-3pt}
\label{fig:arch}
\end{figure}

We describe how data swapping in \ours extends the current workflow of episodic memory (EM) in Figure~\ref{fig:arch}. 
For ease of illustration, we assume that the input stream data is organized by
consecutive tasks but CL learners do not necessarily rely on boundaries
between tasks to perform training and update EM. There are three common
stages involved in traditional EM methods, which proceed in order: \textbf{data
incoming}, \textbf{training}, and \textbf{EM updating}.
This workflow corresponds to many existing methods
including TinyER~\citep{tiny}, CBRS~\citep{pmlr-v119-chrysakis20a}, iCaRL~\citep{icarl},
BiC~\citep{bic}, and DER++~\citep{darker}. 
Then, we add two additional key stages for data swapping: \textbf{storage updating} and \textbf{storage
sample retrieving}.

\begin{myitemize}
\item \textbf{Data incoming \circleb{\scriptsize I}}: The episodic memory
    maintains a subset of samples from previous tasks $\{T_1, \dots, T_{i-1}\}$. When samples
    for a new task $T_i$ arrive, they are first enqueued into a \emph{stream
    buffer} and later exercised for training. Different CL algorithms require
    different amounts of samples to be enqueued for training. The \emph{task-level}
    learning relies on task boundaries, waiting until all $T_i$'s samples
    appear~\citep{icarl, bic}. On the contrary, the \emph{batch-level}
    learning initiates the training stage as soon as a batch of $T_i$'s samples
    in a pre-defined size is available~\citep{tiny, aser, pmlr-v119-chrysakis20a}.

\item \textbf{Training \circleb{\scriptsize T}}: The training combines old
    samples in EM with new samples in a stream buffer to compose a training
    \emph{bundle}. The CL learner organizes the bundle into one or more \emph{mini-batches},
    where each mini-batch is a mixture of old and new samples. The mini-batch size
    and the ratio between the two types of samples within a mini-batch are configured by
    the learning algorithm.
    Typically, several mini-batches are constructed in the
    task-level learning. Learners may go over multiple passes given a bundle,
    trading off computation cost for accuracy.

\item \textbf{EM updating \circleb{\scriptsize M}}: Once the training stage
    is completed, samples in the stream buffer are no longer new and
    represent a past experience, requiring EM to be updated with
    these samples. EM may have enough space to store all of them, but if it
    does not, the CL method applies a sampling strategy like
    \emph{reservoir sampling}~\citep{vitter1985random} and 
    \emph{greedy-balancing sampling}~\citep{gdumb} to select a subset of
    samples from EM as well as from the stream buffer to keep in EM.
    All prior works ``discard'' the samples which
    are not chosen to be kept in EM.

\item \textbf{Storage updating \circleb{\scriptsize S}}: \ours flushes the
    stream data onto the storage before cleaning up the stream buffer for the
    next data incoming step. No loss of information occurs if the free space
    available in the storage is large enough for the stream data. However, if the
    storage is filled due to lack of capacity, we end up having victim
    samples to remove from the storage. In this case, we randomly choose
    samples to evict for each class while keeping the in-storage
    data class-balanced.

\item \textbf{Storage sample retrieving \circleb{\scriptsize R}}: With the
    large number of samples maintained in the storage, data swapping replaces
    in-memory samples with in-storage samples during training to exercise
    abundant information preserved in the storage regarding past experiences.
    \ours collects various useful signals for each in-memory sample used in
    the training stage and determines whether to replace that sample or not.
    This decision is made by our generic \emph{gating function}
    that selects a subset of the samples for replacement with effectively little runtime cost.

\end{myitemize}

Since old samples
for training are drawn directly from $EM$ and a large pool of samples is always kept in the storage $ES$, the training phase is
forced to have a \emph{boundary} of sample selection restricted by
the size of $EM$.
The continual learning with data swapping that optimizes model parameters
$\theta$ for old and new samples $x$ (and corresponding labels $y$) can hence be formulated
as follows: \vspace{-0.5cm}

\begin{equation}
    \mathop{\mathrm{argmin}}_{\theta}\sum_{task\ id=1}^{i}E_{(x, y)\sim ES \cup T_i}[L(f(x,\theta), y)],\text{where}\ (x,y) \in EM.
    \label{eq:em_formula}
\end{equation}

\subsection{Minimizing Delay to Continual Model Training}
\label{sec:SystemDesign}

The primary objective in our proposed design is hiding performance interference caused by the data swapping so that \ours incurs low latency during training. 
To that end, we propose two techniques that encompass in-storage sample retrieval and EM updating stages.

\begin{wrapfigure}{l}{0.45\linewidth}
  \centering
  \vspace{-\intextsep}
  \hspace*{-.75\columnsep}
  \includegraphics[width=\linewidth]{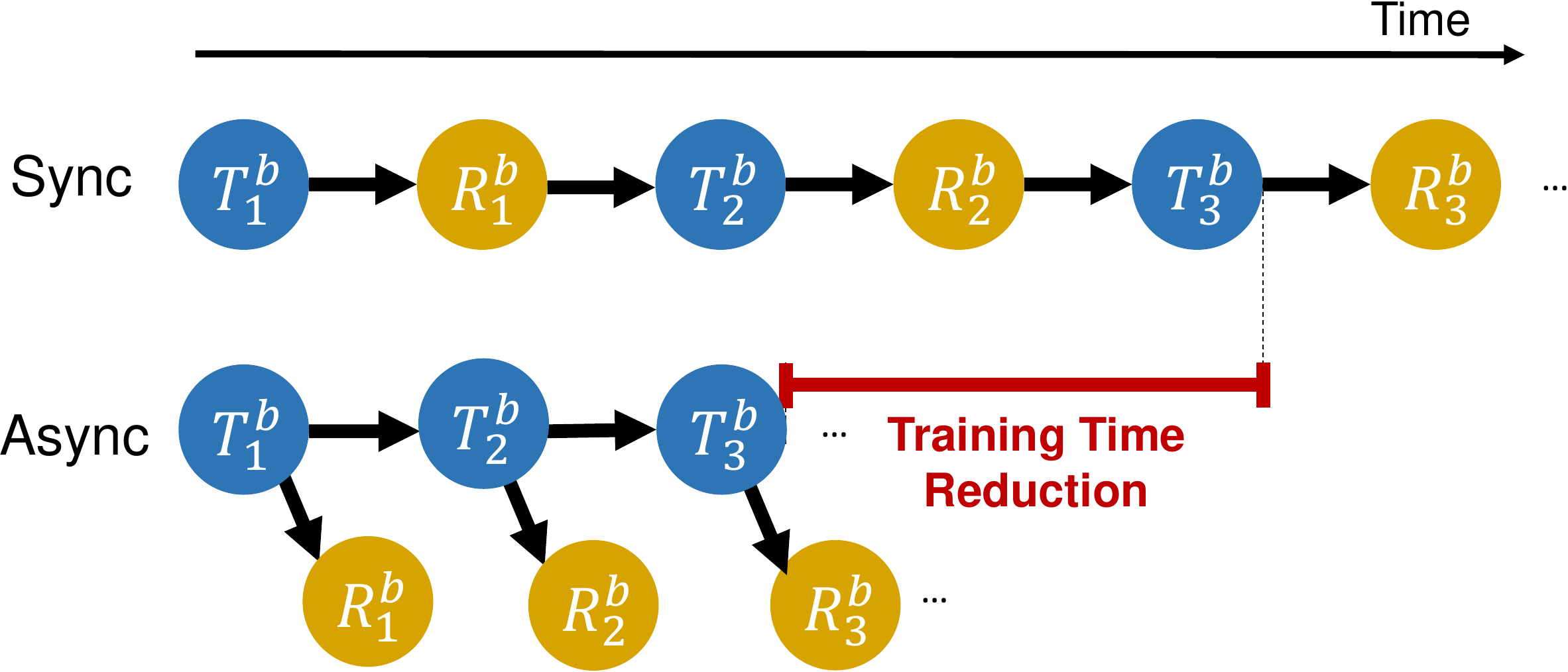}
  \caption{\label{fig:delay_op} Training time reduction with async sample retrieval.}
  \vspace{-3pt}
\end{wrapfigure}

\Paragraph{Asynchronous sample retrieval.} Similar to the conventional learning practice,
\ours maintains fetch workers performing data decoding and
augmentation. As shown in Figure~\ref{fig:arch},
\ours has an additional \emph{swap worker} dedicated to
deciding in-memory samples to evict and issuing I/O requests to bring new
samples from storage into memory.
In the CL workflow, the data retrieval stage $R$ has
dependency on the training stage $T$ since training triggers the replacement
of an in-memory sample when it is used as training input. To illustrate, we
assume that the system has a single fetch worker to pre-process the training
input bundle and creates $N$ mini-batch from the bundle -- this pre-processing
is incurred every time a sample is fetched for training.
The swap worker identifies samples in EM to be replaced from mini-batch $i$
training ($T^b_i$) and then issues I/O requests to retrieve other samples
from storage ($R^b_i$).
If we want to allow the next mini-batch training to exercise EM completely
refreshed with the replaced samples, executions of $T^b$ and $R^b$ by
definition must be serialized such that we have a sequence of
${T^b_1}\rightarrow{R^b_1}\rightarrow{T^b_2}\rightarrow{R^b_2}\rightarrow{T^b_3}\rightarrow{R^b_3}$,
as shown in Figure~\ref{fig:delay_op} (Sync).
However, committing to such strict serialized executions slows down training
speed significantly, \eg, the second mini-batch training $T^b_2$ can start
only after finishing up ${T^b_1}\rightarrow{R^b_1}$, which takes much longer
time than ${T^b_1}$ with no retrieval of storage data as done in the
traditional EM design. To prevent this performance degradation, \ours adopts
\emph{asynchronous sample retrieval} that runs the retrieval step in parallel
with the subsequent training steps. By the asynchronous method, we keep the
minimum possible dependency as shown in Figure~\ref{fig:delay_op} (Async),
with an arbitrary $R^b_i$ not necessarily happened before $T^b_{i+1}$.
Apparently, this design choice poses a delay on applying in-storage samples
to EM, making it possible for the next training steps to access some samples
undergoing replacement. However, we found such accesses do not frequently
occur, and the delay does not nullify the benefit that comes
from data swapping.

In addition, when the swap worker retrieves in-storage samples and writes on
memory, it may interfere with fetch workers that attempt to read samples from
it for pre-processing. To mitigate such interference, \ours could opt for EM
partitioning to parallelize read/write operations over independent
partitions. With EM partitioning, only those operations that access the same
partition need coordination, achieving concurrency against operations that
access other partitions.

\subsection{Data Swapping Policy by a Gate Function}
\label{sec:Algorithms}

The gate function in Figure~\ref{fig:arch} is a core component in \ours for adjusting I/O traffic.
The gate, as guided by its decision logic, allows us to select a
certain portion of samples to swap out from those EM samples drawn in the
training stage. Having this control knob is of big practical importance
as the maximum sustainable I/O traffic differs considerably among devices due
to their in-use storage mediums with different characteristics (\eg,
high-bandwidth flash drive \emph{vs} low-bandwidth magnetic drive). At the same
time, the gate is required to be effective with such partial data swapping
in terms of accuracy in the subsequent training steps.

To facilitate this, we propose a
\emph{sample scoring} method that ranks the samples in the same mini-batch so that
the training algorithm can decide at which point along the
continuum of the ranks we can separate samples to swap from other samples to keep further in memory.

\Paragraph{Score-based replacement.}
The score quantifies the relative importance of a trained sample to keep in EM with respect to other samples in the same mini-batch.
Intuitively, a higher score means that the sample is in a higher rank, so is better ``not'' to be replaced if we need to reduce I/O traffic and vice versa.
To this end, we define the gate function $\sigma_i$ for $i^\text{th}$ sample, $x_i$, as $\sigma_i = \mathbbm{1}(s(x_i) > \tau)$,
where $s(x_i)$ is a scoring function and $\tau$ is a scoring threshold, with both $s(x_i)$ and $\tau$ between $0$ and $1$. 
The threshold is determined by the proportion of the samples that we want to replace from the EM with samples in storage with the consideration of computational efficiency. It allows data swapping to match with I/O bandwidth available on the storage medium, and prevents the system from over-subscribing the bandwidth leading to I/O back-pressure and increased queueing time or under-subscribing the bandwidth leaving storage data exploited less opportunistically.

\Paragraph{Policies.} 
We design several swapping policies driven by the sample scoring method in the context of CL with data swapping for the first time. Specifically, we propose the following three policies:

\underline{(1) Random} selects random samples to replace from EM. Its scoring function assigns 0 to the $\tau$ proportion of the samples randomly selected from a mini-batch while assigning 1 to the other samples.

\underline{(2) Entropy} collects two useful signals for each sample produced
during training: prediction correctness and the associated entropy. This policy prefers to replace the
samples that are correctly predicted because these samples may not be much
beneficial to improve the model in the near future. Furthermore, in this
group of samples, if any specific sample has a lower entropy value than
the other samples, the prediction confidence is relatively
stronger, making it a better candidate for replacement. By
contrast, for the samples that are incorrectly predicted, this policy prefers to
``not'' replace the samples that exhibit lower entropy, \ie, incorrect prediction with stronger confidence, 
since they may take longer to be predicted correctly. Thus, the scoring function $s(x_i)$ 
with a model $f(\cdot)$ is defined as:

\begin{equation}
    s(x_i) = \frac{1}{U} \left[ g(x_i)(H(f(x_i))) + (1 - g(x_i)) \left(U - H(f(x_i)) \right) \right],
    \label{eq:score}
\end{equation}

where $g(x_i) = \mathbbm{1}(f(x_i)=y_i)$, $H(\cdot)$ is an entropy function, and $U$ is the maximum entropy value.

\underline{(3) Dynamic} combines Random and Entropy to perform
the first half of training passes given a bundle with Random and the next
half of the passes with Entropy. This policy is motivated by curriculum learning~\citep{curr},
which gradually focuses on training harder samples as time elapses.

It is indeed possible to come up with a number of replacement policies, for which this paper introduces a few concrete examples. 
Regardless, designing the gate logic with more effective replacement policies is a promising research direction that we want to further explore in \ours.

\section{Experiments}
\label{sec:Experiments}

As \ours is broadly applicable to a variety of EM-based CL methods, we
compare the performance with and without \ours in the methods of their own
setups. We select seven methods as shown in
Table~\ref{tab:punchline}, to cover several aspects discussed in
Section~\ref{sec:Approach} such as bundle boundary of learning (\ie,
task-level \emph{vs} batch-level)
and number of passes taken per bundle.
We discuss detailed reproducible settings in Section~\ref{sec:Reproducibility}. 
For evaluation, we implement \ours in PyTorch 1.7.1 as a working prototype.

\Paragraph{Datasets and metrics.} 
Datasets include \textbf{CIFAR subset}---CIFAR10 (\textbf{C10}) and CIFAR100 (\textbf{C100})---, \textbf{ImageNet subset}---ImageNet-100 (\textbf{I100}), Mini-ImageNet (100 classes) (\textbf{MI100}), and Tiny-ImageNet (200 classes) (\textbf{TI200})---, and \textbf{ImageNet-1000}.
We use two
popular metrics, the \textbf{final accuracy} and the \textbf{final forgetting}~\citep{rwalk} averaged over classes, to reflect the performance
of continual learning. Except for ImageNet-1000 that represents a significantly large-scale training,
the results are averaged over five runs while each
method assigns an equal of classes to each task. We also measure \textbf{training
speed} measured from the time the training stage receives a bundle
to the time it completes training the bundle.

\Paragraph{Baselines and architectures.}
On top of each CL method, we vary the amount of data swapping to study the
effectiveness of \ours in detail. Unless otherwise stated, \textbf{\ours-N}
means that our swap worker is configured to replace N\% of EM samples drawn
by the training stage.
All experiments are based on either ResNet or DenseNet neural networks, with
all using the SGD optimizer as suggested by the original works, 
and use the entropy-based data swapping policy (\ie, \textbf{Entropy}) by default. 

\subsection{Results}
\label{sec:Results}

\begin{table}[t]
\centering
\scriptsize
\resizebox{0.99\textwidth}{!}{
\begin{tabular}{lclccccccc}
\toprule

& \multirow{3.5}{*}{{Metric}} & {{Method}} & ER & iCaRL & TinyER & BiC & GDumb & DER++ & RM\\
\cmidrule(l{5pt}r{5pt}){3-3} \cmidrule(l{5pt}r{5pt}){4-4} \cmidrule(l{5pt}r{5pt}){5-5} \cmidrule(l{5pt}r{5pt}){6-6}  \cmidrule(l{5pt}r{5pt}){7-7}  \cmidrule(l{5pt}r{5pt}){8-8}  \cmidrule(l{5pt}r{5pt}){9-9}  \cmidrule(l{5pt}r{5pt}){10-10}
& & (EM Sizes) & (2000/2000) & (2000/2000) & (300/500) & (2000/2000) & (500/4500) & (500/500) & (500/4500)\\
& & (Datasets) & (C100/I100) & (C100/I100) & (C100/MI100) & (C100/I100) & (C10/TI200) & (C10/TI200) & (C10/TI200)\\

\midrule

\multirow{8}{*}{\rotatebox[origin=c]{90}{CIFAR Subset}}
& \multirow{3.5}{*}{{Acc. ($\uparrow$)}}
& Original             & 33.66$\pm$0.42 & 47.04$\pm$0.25 & 54.11$\pm$2.52 & 48.62$\pm$0.37 & 47.29$\pm$0.54 & 72.17$\pm$1.11 & 52.15$\pm$1.42\\
\cdashlinelr{3-10}
& & \textbf{\ours-50}   & 54.00$\pm$0.49 & 48.39$\pm$0.41 & 59.99$\pm$2.12 & 62.40$\pm$0.40 & 52.64$\pm$1.64 & 90.05$\pm$0.38 & 66.66$\pm$0.75\\
& & \textbf{\ours-100}     & \textbf{54.93$\pm$0.46} & \textbf{49.48$\pm$0.53} & \textbf{61.83$\pm$2.50} & \textbf{63.07$\pm$0.48} & \textbf{53.72$\pm$1.06} & \textbf{90.58$\pm$0.30} & \textbf{67.81$\pm$0.51}\\
\cmidrule{2-10}

& \multirow{3.5}{*}{{Fgt. ($\downarrow$)}}
& Original             & 57.60$\pm$0.54 & 22.38$\pm$0.30 & 16.04$\pm$2.91 & 20.21$\pm$0.65 & 22.14$\pm$0.83 & 24.45$\pm$1.81 & 23.03$\pm$2.86\\
\cdashlinelr{3-10}
& & \textbf{\ours-50}   & \textbf{32.13$\pm$0.72} & 17.18$\pm$0.53 & 11.9$\pm$1.13 & 13.40$\pm$0.86 & \textbf{20.92$\pm$1.63} & 3.38$\pm$0.22 & \textbf{9.63$\pm$0.40}\\
& & \textbf{\ours-100}     & 32.28$\pm$0.55 & \textbf{16.73$\pm$0.65} & \textbf{9.44$\pm$1.96} & \textbf{12.16$\pm$0.69} & 21.18$\pm$1.49 & \textbf{2.78$\pm$0.55} & 11.03$\pm$2.28\\
\midrule

\multirow{8}{*}{\rotatebox[origin=c]{90}{ImageNet Subset}}
& \multirow{3.5}{*}{{Acc. ($\uparrow$)}}
& Original             & 65.62$\pm$1.48 & 79.95$\pm$1.12 & 56.91$\pm$2.32 & 83.05$\pm$0.66 & 42.25$\pm$0.44 & 19.38$\pm$1.41 & 21.84$\pm$0.54\\
\cdashlinelr{3-10}
& & \textbf{\ours-50}   & 82.21$\pm$0.43 & 79.15$\pm$0.40 & 65.43$\pm$2.83 & \textbf{88.03$\pm$0.39} & 55.67$\pm$0.21 & \textbf{47.74$\pm$0.84} & 45.49$\pm$0.32\\
& & \textbf{\ours-100}     & \textbf{83.56$\pm$0.36} & \textbf{80.15$\pm$1.07} & \textbf{68.70$\pm$0.59} & 87.63$\pm$0.34 & \textbf{56.11$\pm$0.47} & 45.12$\pm$1.39 & \textbf{46.70$\pm$0.16}\\
\cmidrule{2-10}

& \multirow{3.5}{*}{{Fgt. ($\downarrow$)}}
& Original             & 55.30$\pm$0.28 & 17.10$\pm$0.68 & 15.30$\pm$2.29 & 15.81$\pm$0.34 & 11.06$\pm$0.75 & 66.83$\pm$0.93 & 16.89$\pm$0.51\\
\cdashlinelr{3-10}
& & \textbf{\ours-50}   & 40.12$\pm$0.55 & 14.65$\pm$0.27 & \textbf{7.64$\pm$4.08} & \textbf{7.96$\pm$0.58} & 9.47$\pm$0.23 & \textbf{20.69$\pm$1.52} & 9.43$\pm$0.39\\
& & \textbf{\ours-100}     & \textbf{39.25$\pm$0.29} & \textbf{14.28$\pm$0.20} & 9.76$\pm$2.48 & 8.27$\pm$0.28 & \textbf{8.01$\pm$1.84} & 23.81$\pm$1.68 & \textbf{8.81$\pm$0.21}\\
\midrule

\multirow{8}{*}{\rotatebox[origin=c]{90}{ImageNet-1000}}
& \multirow{3.5}{*}{{Acc. ($\uparrow$)}}
& Original             & 40.47 & \textbf{56.69} & 46.41 & 77.04 & 38.14 & 12.11 & 24.08\\
\cdashlinelr{3-10}
& & \textbf{\ours-50}   & 42.64 & 55.72 & 48.93 & 80.66 & 48.86 & 35.90 & 43.69\\
& & \textbf{\ours-100}   & \textbf{44.21} & 55.45 & \textbf{50.30} & \textbf{80.84} & \textbf{49.56} & \textbf{36.89} & \textbf{44.36}\\
\cmidrule{2-10}

& \multirow{3.5}{*}{{Fgt. ($\downarrow$)}}
& Original             & 70.67 & 39.41 & 10.37 & 21.38 & 31.56 & 54.78 & 12.3\\
\cdashlinelr{3-10}
& & \textbf{\ours-50}   & 65.60 & 40.37 & 8.79 & 18.81 & 25.12 & 24.31 & 7.31\\
& & \textbf{\ours-100}   & \textbf{64.79} & \textbf{39.02} & \textbf{7.20} & \textbf{18.30} & \textbf{23.86} & \textbf{21.77} & \textbf{7.16}\\

\bottomrule

\end{tabular}
}
\vspace{-1em}
\caption{Accuracy and Forgetting of EM methods with and without proposed designs of \ours-50 (50\% swapping) and \ours-100 (100\% swapping). For each method, the two parenthesises show EM sizes and datasets used for the CIFAR subset and the ImageNet subset. For the ImageNet-1000, all methods use the EM size of 20000 and runs once due to significant training time. ($\uparrow$) higher is better, ($\downarrow$) lower is better. Please refer to the `datasets and metrics' paragraph for dataset name abbreviations.}
\label{tab:punchline}
\vspace{-3pt}
\end{table}

We compare existing methods with two \ours versions,
\textbf{\ours-50} that performs partial swapping for
a half of the data and \textbf{\ours-100} that performs full swapping.
Table~\ref{tab:punchline}
presents the performance in terms of the top-1 accuracy (\textbf{Acc.}) and 
the forgetting score (\textbf{Fgt.}), except for ER, iCarL, and
BiC that measure the top-5 accuracy for the ImageNet subset as done in the
original works. 

First, \ours-100 improves the accuracy remarkably over almost all of the methods
under consideration, advancing the state-of-the-art performances for CIFAR and ImageNet datasets.
The results clearly show the effectiveness of using the storage device in large capacity
to allow CL to exploit abundant information of the previous tasks.
Among the seven methods, \ours-100 delivers relatively larger accuracy gains
for BiC~\citep{bic}, GDumb~\citep{gdumb}, DER++~\citep{darker}, and RM~\citep{rainbow} that take multi-passes on each training
input. We believe that as long as old samples in EM are exercised more
frequently for a new bundle to train (\ie, new samples plus old samples), data swapping can
subsequently bring in more diverse samples from storage to take advantage of
them. Regardless, although TinyER~\citep{tiny} is designed to take a
single pass over new samples and thus exercise EM less aggressively, as applied
with our techniques, it improves the accuracy by 7.72\%, 11.79\%, and 3.89\% for CIFAR-100, Mini-ImageNet, and ImageNet-1000, respectively.

In comparison to \ours-100, \ours-50 obtains slightly lower accuracy across the
models. We argue that such a small sacrifice in accuracy is indeed
worthwhile when storage I/O bandwidth is the primary constraint.
In \ours-50, with 50\% lower I/O traffic caused by data swapping,
the accuracy as compared to \ours-100 diminishes only by 1\%, 0.6\%, and 0.7\%
on average for CIFAR subset, ImageNet subset, and ImageNet-1000, respectively, providing an ability to
trade-off small accuracy loss for substantial I/O traffic reduction.
Similarly to the accuracy, our data swapping approaches
considerably reduce forgetting scores over the majority of the original methods.
Perhaps, one method that shows
less promising results in Table~\ref{tab:punchline} would be iCaRL~\citep{icarl},
where \ours makes the accuracy occasionally worse. 

From the in-depth investigation of iCaRL in Appendix~\ref{sec:distill_detail}, we observe that using data swapping and knowledge distillation at the same time cannot not deliver great accuracy. That is, as knowledge distillation may not be much compatible with data swapping, we
revisit distillation-based CL methods (\ie, iCaRL, BiC, and DER++) when they
are used along with data swapping in detail.

\begin{figure}[t]
\center
\includegraphics[width=\linewidth]{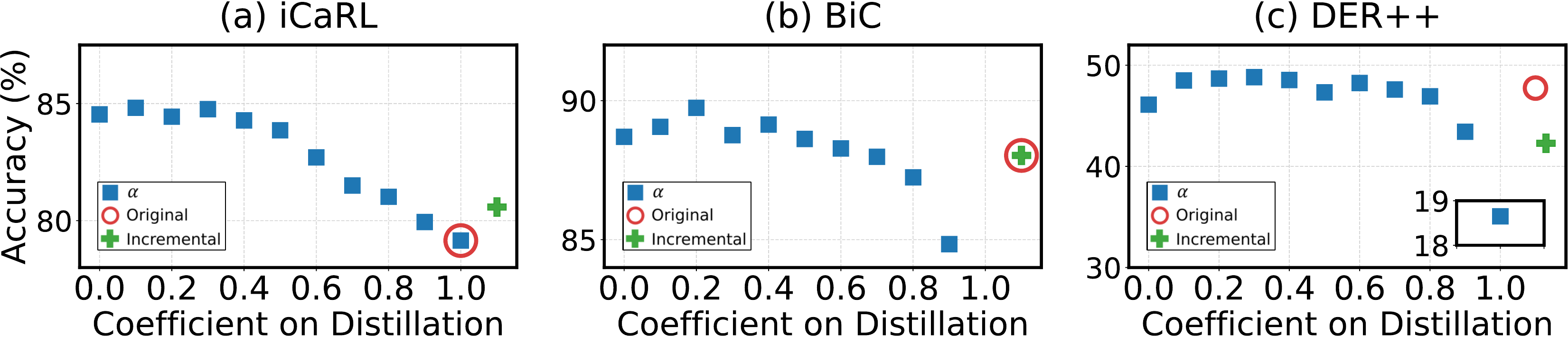}
\caption{Accuracy of distillation-based methods with \ours-50 on ImageNet subset while varying coefficient ($\alpha$) values on the distillation loss in calculating training loss.
}
\label{fig:distill}
\vspace{-3pt}
\end{figure}

\Paragraph{Knowledge distillation on \ours.} 
Note that the ways to distill the knowledge of old data in iCaRL, BiC, and
DER++ are all different (see Appendix~\ref{sec:distill_detail}). Briefly speaking, in
calculating loss for old data, iCaRL uses only soft labels obtained from an
old classifier, whereas BiC and DER++ use both hard labels
(\ie, ground truth) as well as soft labels. To investigate the effect of using these two types of
loss, we first modify the loss function of iCaRL similarly to that of BiC,
\ie, $\alpha \times soft\ label\ loss + (1-\alpha) \times hard\ label\
loss$, and then show accuracy over varying $\alpha$ values for all three distillation-based methods
in Figure~\ref{fig:distill}. For each method, we also include accuracy when $\alpha$
increases incrementally over time as done in BiC.

\begin{table}[t]
\centering
\scriptsize
\resizebox{0.85\textwidth}{!}{
\begin{tabular}{lccccccc}
    \toprule
    Method & ER & iCaRL & TinyER & BiC & GDumb & DER++ & RM\\
    \midrule
    \textbf{Random}   & 82.07$\pm$0.28 & \textbf{79.31$\pm$0.30} & 65.04$\pm$2.75 & 87.81$\pm$0.87 & 55.57$\pm$0.28 & 47.37$\pm$0.91 & 45.46$\pm$0.46\\
    \textbf{Entropy}   & 82.21$\pm$0.43 & 79.15$\pm$0.40 & \textbf{65.43$\pm$2.83} & \textbf{88.03$\pm$0.93} & \textbf{55.67$\pm$0.21} & \textbf{47.74$\pm$0.84} & 45.49$\pm$0.32\\
    \textbf{Dynamic}  & \textbf{82.94$\pm$0.43} & 79.14$\pm$0.31 & 65.04$\pm$2.75 & 87.79$\pm$0.88 & 55.51$\pm$0.74 & 47.53$\pm$2.09 & \textbf{45.80$\pm$0.25}\\
    \bottomrule
\end{tabular}
}
\caption{Comparison of accuracy for data swapping policies for \ours-50 on ImageNet subset.
}
\label{tab:policies}
\vspace{-3pt}
\end{table}

The results show that distillation-based methods with CarM significantly
improve accuracy when the $\alpha$ is very small. For iCaRL, compared to
$\alpha = 1.0$ (\ie, no hard label loss as iCaRL does), we obtain 5.4 higher
accuracy when $\alpha = 0.0$ (\ie, no distillation) and 5.7 higher accuracy
when $\alpha = 0.1$, which is the best result. Similarly, for BiC and DER++
with CarM, we found that the coefficient $\alpha$ to be applied on the soft
label loss does not necessarily be high (\ie, iCaRL) or managed complicatedly (\ie, BiC) to achieve higher accuracy. Please refer to Figure~\ref{fig:distill_cifar} for the CIFAR subset results.

Our best interpretation for the reason behind is as follows.
The key assumption of knowledge distillation is that once the model is trained with a new task,
the knowledge newly learned is supposed to generalize the task well and can
be effectively transferable to subsequent task training. However, if the
model is not sufficiently generalized for old tasks, using distillation
losses extensively might be adverse---Data swapping attempts to correct
decision boundaries driven by abundant in-storage samples to further
generalize old tasks, but interfered by the knowledge distilled by the old
models.

\Paragraph{Comparison for data swapping policies.}
We compare the performance of the three data swapping policies proposed in Section~\ref{sec:Algorithms} under \ours-50.
As shown in Table~\ref{tab:policies}, both Entropy and Dynamic outperform Random by 0.16\% on average for ImageNet subset (see Table~\ref{tab:policies_cifar} for CIFAR subset). 
We highlight that our major contribution for gating mechanism is computational efficiency while matching with I/O bandwidth available on the storage medium, and the primary objective of exploring data swapping policies is establishing a good baseline for the gating mechanism. 
In this regard, we found that all three policies can serve as good baselines.

\begin{table}[t]
\centering
\scriptsize
\resizebox{0.99\textwidth}{!}{
\begin{tabular}{lccccccc}
    \toprule
    Method & ER & iCaRL & TinyER & BiC & GDumb & DER++ & RM\\
    \midrule
    \textbf{Async}   & +0.7\%/+2.9\% & +0.5\%/-2.7\% & +3.3\%/+5.5\% & -0.3\%/+1.7\% & +3.3\%/+0.1\% & +0.3\%/+2.4\% & +2.3\%/-0.9\%\\
    \textbf{Sync}  & +30.2\%/+21.3\% & +70.5\%/+62.0\% & +11.3\%/+10.7\% & +20.0\%/+33.8\% & +52.1\%/+7.6\% & +71.6\%/+38.8\% & +25.9\%/+2.6\%\\
    \bottomrule
\end{tabular}
}
\caption{Training speed with data swapping for \ours-50. (+) training time increases, (-) training time decreases.
Each field includes the results as a pair for the CIFAR subset and the ImageNet subset.}
\label{tab:speed}
\vspace{-3pt}
\end{table}

\Paragraph{Impact on training speed.} 
Delay optimization techniques in Section~\ref{sec:SystemDesign} are
intended to incur insignificant delay on training.
To confirm this, we examine how training speed in \ours-50 changes over the original memory-only methods,
measured as the percentage of wall-clock time (\ie,
actual time taken) increase as applied with asynchronous (\textbf{Async}) \emph{vs} synchronous sample retrieval (\textbf{Sync}).
To consider the most challenging scenario,
we make data entered into the stream buffer at a rate enough to keep training
always busy with new mini-batches. As shown in Table~\ref{tab:speed},
regardless of EM methods, the asynchronous version of \ours does not dramatically affect
training speeds for both CIFAR and ImageNet subsets.
By contrast, the synchronous version slows down training time up to 71.6\% for CIFAR subset and 62.0\% for ImageNet subset. 
Regardless of the version in use, in-memory samples undergoing data swapping are rarely drawn in the subsequent training steps since the size of an episodic memory size is typically much larger than the size of a training batch. Therefore, no difference in accuracy is observed between the two version.

\begin{figure}[!t]
\center
\includegraphics[width=\linewidth]{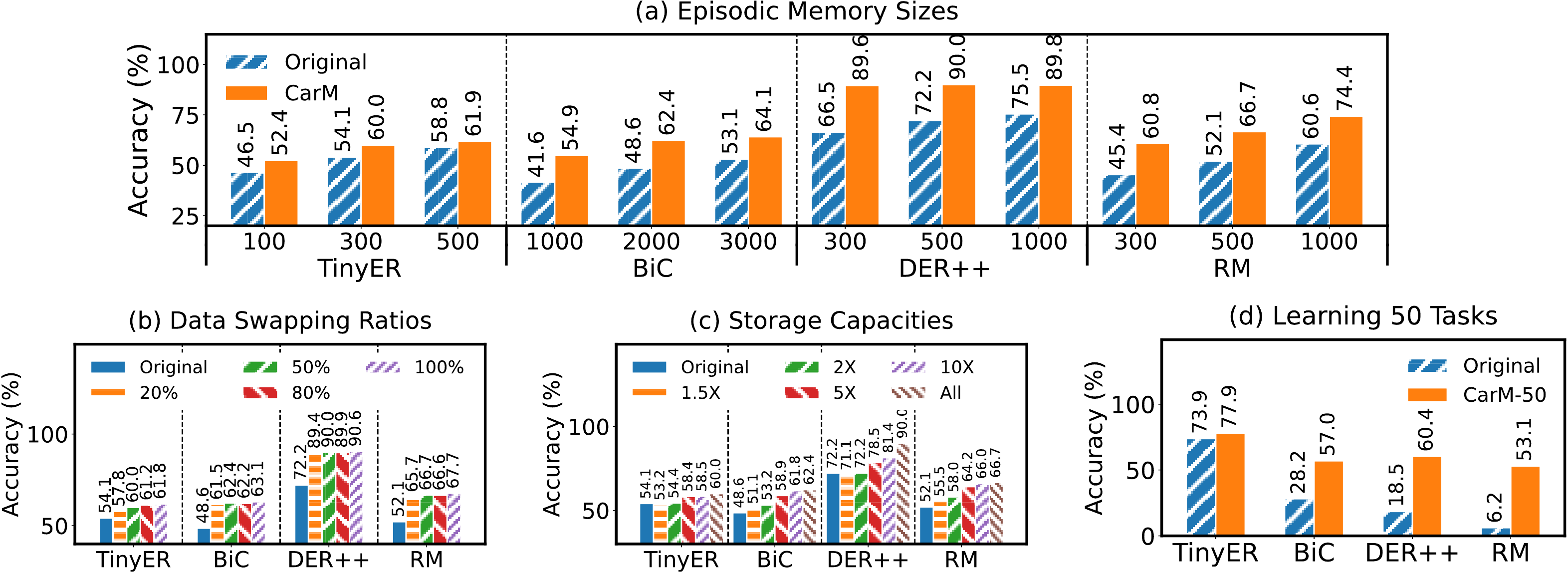}
\caption{Accuracy over varying (a) EM sizes, (b) data swapping ratios, and (c) storage capacities, and for (d) learning 50 tasks. For (a), larger and smaller EMs are chosen by referring to Table~\ref{tab:punchline}.}
\label{fig:ablation}
\vspace{-3pt}
\end{figure}

\subsection{Ablation Study}
\label{sec:Ablation}

We present an ablation study using four methods (TinyER, BiC, DER++, and RM) that represent the state-of-the-art in
each type of EM methodologies, using the CIFAR subset.

\Paragraph{Size of EM.}
To confirm performance benefits over using different memory sizes, we empirically evaluate \ours-50
over varying EM sizes and show the average accuracy in Figure~\ref{fig:ablation}(a).
In all cases, \ours-50 outperforms the existing methods, with BiC, DER++, and RM having
relatively higher accuracy increases. Moreover, we observe that data swapping delivers
better accuracy over conventional memory-only approaches
using much smaller memory. For example,
\ours-50 with DER++ on EM size 300 shows higher accuracy than pure DER++ on EM size 1000,
and \ours-50 with TinyER on EM size 300 shows higher accuracy than pure TinyEM on EM size 500.
Therefore, it turns out that data swapping could help reduce the EM size without hurting the accuracy of existing methods in both the
multi-pass method and single-pass method.

\Paragraph{Data swapping ratio.} 
We present results with different swapping
ratios to show that our gate model indeed brings out meaningful benefits over
using different I/O bandwidths. To that end, Figure~\ref{fig:ablation}(b) shows
the change in accuracy when our gating policy decreases the swapping ratio
down to 20\% (\ours-20) or increases it up to 80\% (\ours-80). Obviously, at
\ours-80 in high swapping ratio, the accuracy across the four EM methods gets
very close to the accuracy obtained in full swapping. A surprising result is
that even at \ours-20 in 20\% data swapping, the accuracy is very comparable
to when we allow higher data swapping ratios. The results indicate that our
method would be effective even when applied to the system with low-bandwidth
storage.

\Paragraph{Size of storage.}
As local storage cannot store
all the past data, the system must discard some old
samples once the storage is fully occupied.
Figure~~\ref{fig:ablation}(c) shows accuracy degradation in \ours-50 when storage
capacity is limited to 1.5--10$\times$ of the EM size. The results show that 
data swapping
improves performance over traditional approaches even with using 50\% larger capacity for the storage.

\Paragraph{Large number of tasks.}
One pressing issue on CL is learning a
large number of tasks as it is required to keep the knowledge learned in the
remote past. To evaluate this aspect, we split CIFAR-100 (100 classes) into
50 tasks and run with the four methods. As Figure~\ref{fig:ablation}(d)
shows, \ours significantly outperforms the baselines, showing the potential
for long-term continual model training.

\section{Conclusion}
\label{sec:Conclusion}

We alleviate catastrophic forgetting by integrating traditional episodic memory-based continual learning methods with
device-internal data storage, named \ours. We design data swapping strategies to improve
model accuracy by dynamically utilizing a large amount of the past data
available in the storage. Our swapping mechanism addresses the cumbersome
performance hurdle incurred by slow storage access, and hence continual model
training is not dramatically affected by data transfers between memory and
storage. We show the effectiveness of \ours using seven well-known methods on
standard datasets, over varying memory sizes, storage sizes, and data
swapping ratios.



\section*{Reproducibility Statement}
\label{sec:Reproducibility}


We take the reproducibility of the research very seriously. Appendix hence includes detailed information necessary for reproducing
all the experiments performed in this work, as follows:

\begin{myitemize}
\item Appendix~\ref{sec:Impledetails} describes the implementation details of building \ours.
\item Appendix~\ref{sec:Datasets} specifies dataset information used in the experiments (\eg, the number of tasks and the number of classes per task).
\item Appendix~\ref{sec:Expdetails} provides experimental details (\eg, metrics and hyper-parameters).
\item Appendix~\ref{sec:Machines} presents detailed specification of machines (\eg, GPU model) used in the experiments.
\end{myitemize}
   
Our source code is available at \url{https://github.com/supersoob/CarM}, where we include running environments and configuration files for all the experiments that make it possible to reproduce the results reported in this paper with minimal effort.


\section*{Ethics Statement}

All continual learning (CL) methods including the proposed one would adapt and extend the already trained AI model to recognize better with the streamed data.
The CL methods will expedite the deployment of AI systems to help humans by its versatility of adapting to a new environment out of the factory or research labs.
As all CL methods, however, would suffer from adversarial streamed data as well as data bias, which may cause ethnic, gender or biased gender issues, the proposed method would not be an exception.
Although the proposed method has \emph{no intention} to allow such problematic cases, the method may be exposed to such threats.
Relentless efforts should be made to develop mechanisms to prevent such usage cases in order to make the continuously updating machine learning models safer and enjoyable to be used by humans.

{\bibliographystyle{iclr2022_conference}
\bibliography{iclr2022_conference}

\begin{thebibliography}{54}
\providecommand{\natexlab}[1]{#1}
\providecommand{\url}[1]{\texttt{#1}}
\expandafter\ifx\csname urlstyle\endcsname\relax
  \providecommand{\doi}[1]{doi: #1}\else
  \providecommand{\doi}{doi: \begingroup \urlstyle{rm}\Url}\fi

\bibitem[asy()]{asyncio}
Python asynchronous i/o.
\newblock \url{https://docs.python.org/3.8/library/asyncio.html}.

\bibitem[man()]{manager}
Python manager.
\newblock
  \url{https://docs.python.org/3.8/library/multiprocessing.html#managers}.

\bibitem[pyt()]{pythonmulti}
Python multiprocessing.
\newblock \url{https://docs.python.org/3.8/library/multiprocessing.html}.

\bibitem[sha()]{sharedmem}
Python shared memory.
\newblock \url{
  https://docs.python.org/3.8/library/multiprocessing.shared_memory.html }.

\bibitem[Aljundi et~al.(2019)Aljundi, Lin, Goujaud, and Bengio]{gss}
Rahaf Aljundi, Min Lin, Baptiste Goujaud, and Yoshua Bengio.
\newblock Gradient based sample selection for online continual learning.
\newblock In \emph{NIPS}, 2019.

\bibitem[Ananthanarayanan et~al.(2012)Ananthanarayanan, Ghodsi, Warfield,
  Borthakur, Kandula, Shenker, and Stoica]{pacman}
Ganesh Ananthanarayanan, Ali Ghodsi, Andrew Warfield, Dhruba Borthakur,
  Srikanth Kandula, Scott Shenker, and Ion Stoica.
\newblock Pacman: Coordinated memory caching for parallel jobs.
\newblock In \emph{USENIX NSDI}, pp.\  267--280, 2012.

\bibitem[Bang et~al.(2021)Bang, Kim, Yoo, Ha, and Choi]{rainbow}
Jihwan Bang, Heesu Kim, YoungJoon Yoo, Jung-Woo Ha, and Jonghyun Choi.
\newblock Rainbow memory: Continual learning with a memory of diverse samples.
\newblock In \emph{CVPR}, 2021.

\bibitem[Bengio et~al.(2009)Bengio, Louradour, Collobert, and Weston]{curr}
Yoshua Bengio, J{\'e}r\^{o}me Louradour, Ronan Collobert, and Jason Weston.
\newblock Curriculum learning.
\newblock In \emph{ICML}, 2009.

\bibitem[Borsos et~al.(2020)Borsos, Mutn{\`y}, and
  Krause]{Borsos2020CoresetsVB}
Zal{\'a}n Borsos, Mojm{\'\i}r Mutn{\`y}, and Andreas Krause.
\newblock Coresets via bilevel optimization for continual learning and
  streaming.
\newblock In \emph{NIPS}, 2020.

\bibitem[Buzzega et~al.(2020)Buzzega, Boschini, Porrello, Abati, and
  CALDERARA]{darker}
Pietro Buzzega, Matteo Boschini, Angelo Porrello, Davide Abati, and SIMONE
  CALDERARA.
\newblock Dark experience for general continual learning: a strong, simple
  baseline.
\newblock In \emph{NIPS}, 2020.

\bibitem[Castro et~al.(2018)Castro, Marin-Jimenez, Guil, Schmid, and
  Alahari]{castro2018eccv}
Francisco~M. Castro, Manuel~J. Marin-Jimenez, Nicolas Guil, Cordelia Schmid,
  and Karteek Alahari.
\newblock End-to-end incremental learning.
\newblock In \emph{ECCV}, 2018.

\bibitem[Chaudhry et~al.(2018)Chaudhry, Dokania, Ajanthan, and Torr]{rwalk}
Arslan Chaudhry, Puneet~K. Dokania, Thalaiyasingam Ajanthan, and Philip H.~S.
  Torr.
\newblock Riemannian walk for incremental learning: Understanding forgetting
  and intransigence.
\newblock In \emph{ECCV}, 2018.

\bibitem[Chaudhry et~al.(2019{\natexlab{a}})Chaudhry, Ranzato, Rohrbach, and
  Elhoseiny]{AGEM}
Arslan Chaudhry, Marc’Aurelio Ranzato, Marcus Rohrbach, and Mohamed
  Elhoseiny.
\newblock Efficient lifelong learning with {A-GEM}.
\newblock In \emph{ICLR}, 2019{\natexlab{a}}.

\bibitem[Chaudhry et~al.(2019{\natexlab{b}})Chaudhry, Rohrbach, Elhoseiny,
  Ajanthan, Dokania, Torr, and Ranzato]{tiny}
Arslan Chaudhry, Marcus Rohrbach, Mohamed Elhoseiny, Thalaiyasingam Ajanthan,
  Puneet~K Dokania, Philip~HS Torr, and Marc'Aurelio Ranzato.
\newblock On tiny episodic memories in continual learning.
\newblock \emph{arXiv preprint arXiv:1902.10486}, 2019{\natexlab{b}}.

\bibitem[Chrysakis \& Moens(2020)Chrysakis and Moens]{pmlr-v119-chrysakis20a}
Aristotelis Chrysakis and Marie-Francine Moens.
\newblock Online continual learning from imbalanced data.
\newblock In \emph{ICML}, volume 119, pp.\  1952--1961, 2020.

\bibitem[Cong et~al.(2020)Cong, Zhao, Li, Wang, and Carin]{Cong2020GANMW}
Yulai Cong, Miaoyun Zhao, Jianqiao Li, Sijia Wang, and Lawrence Carin.
\newblock {GAN} memory with no forgetting.
\newblock In \emph{NIPS}, 2020.

\bibitem[Douillard et~al.(2020)Douillard, Cord, Ollion, Robert, and
  Valle]{podnet}
Arthur Douillard, Matthieu Cord, Charles Ollion, Thomas Robert, and Eduardo
  Valle.
\newblock Podnet: Pooled outputs distillation for small-tasks incremental
  learning.
\newblock In \emph{ECCV}, 2020.

\bibitem[Fini et~al.(2020)Fini, Lathuili{\`e}re, Sangineto, Nabi, and
  Ricci]{Fini2020OnlineCL}
Enrico Fini, St{\'e}phane Lathuili{\`e}re, Enver Sangineto, Moin Nabi, and
  Elisa Ricci.
\newblock Online continual learning under extreme memory constraints.
\newblock In \emph{ECCV}, 2020.

\bibitem[Gepperth \& Hammer(2016)Gepperth and Hammer]{gepperthH16}
Alexander Gepperth and Barbara Hammer.
\newblock {Incremental learning algorithms and applications}.
\newblock In \emph{{ESANN}}, 2016.

\bibitem[Hayes et~al.(2020)Hayes, Kafle, Shrestha, Acharya, and
  Kanan]{Hayes2020REMINDYN}
Tyler~L Hayes, Kushal Kafle, Robik Shrestha, Manoj Acharya, and Christopher
  Kanan.
\newblock Remind your neural network to prevent catastrophic forgetting.
\newblock In \emph{ECCV}, 2020.

\bibitem[Hildebrand et~al.(2020)Hildebrand, Khan, Trika, Lowe-Power, and
  Akella]{AutoTM_asplos20}
Mark Hildebrand, Jawad Khan, Sanjeev Trika, Jason Lowe-Power, and Venkatesh
  Akella.
\newblock Autotm: Automatic tensor movement in heterogeneous memory systems
  using integer linear programming.
\newblock In \emph{ASPLOS}, pp.\  875--890, 2020.

\bibitem[Hu et~al.(2019)Hu, Lin, Liu, Tao, Tao, Ma, Zhao, and
  Yan]{hu2018overcoming}
Wenpeng Hu, Zhou Lin, Bing Liu, Chongyang Tao, Zhengwei Tao, Jinwen Ma, Dongyan
  Zhao, and Rui Yan.
\newblock Overcoming catastrophic forgetting via model adaptation.
\newblock In \emph{ICLR}, 2019.

\bibitem[Huang et~al.(2020)Huang, Jin, and Li]{SwapAdvisor_asplos20}
Chien-Chin Huang, Gu~Jin, and Jinyang Li.
\newblock Swapadvisor: Pushing deep learning beyond the gpu memory limit via
  smart swapping.
\newblock In \emph{ASPLOS}, pp.\  1341--1355, 2020.

\bibitem[Jin et~al.(2018)Jin, Liu, Jiang, Ma, Shi, He, and Zhao]{Layrub}
Hai Jin, Bo~Liu, Wenbin Jiang, Yang Ma, Xuanhua Shi, Bingsheng He, and Shaofeng
  Zhao.
\newblock Layer-centric memory reuse and data migration for extreme-scale deep
  learning on many-core architectures.
\newblock In \emph{ACM TACO}, volume~15, pp.\  1--26, 2018.

\bibitem[Jin et~al.(2020)Jin, Du, and Ren]{gmed}
Xisen Jin, Junyi Du, and Xiang Ren.
\newblock Gradient based memory editing for task-free continual learning.
\newblock \emph{arXiv preprint arXiv:2006.15294}, 2020.

\bibitem[Kang et~al.(2020)Kang, Jo, Nam, and Choi]{kangJNC2020confcal}
Dongmin Kang, Yeonsik Jo, Yeongwoo Nam, and Jonghyun Choi.
\newblock Confidence calibration for incremental learning.
\newblock In \emph{IEEE Access}, volume~8, pp.\  126648--126660, 2020.

\bibitem[Kirkpatrick et~al.(2017)Kirkpatrick, Pascanu, Rabinowitz, Veness,
  Desjardins, Rusu, Milan, Quan, Ramalho, Grabska-Barwinska,
  et~al.]{Kirkpatrick2017OvercomingCF}
James Kirkpatrick, Razvan Pascanu, Neil Rabinowitz, Joel Veness, Guillaume
  Desjardins, Andrei~A Rusu, Kieran Milan, John Quan, Tiago Ramalho, Agnieszka
  Grabska-Barwinska, et~al.
\newblock Overcoming catastrophic forgetting in neural networks.
\newblock In \emph{PNAS}, 2017.

\bibitem[Lee et~al.(2017)Lee, Kim, Jun, Ha, and Zhang]{Lee2017OvercomingCF}
Sang-Woo Lee, Jin-Hwa Kim, Jaehyun Jun, Jung-Woo Ha, and Byoung-Tak Zhang.
\newblock Overcoming catastrophic forgetting by incremental moment matching.
\newblock In \emph{NIPS}, 2017.

\bibitem[Li \& Hoiem(2017)Li and Hoiem]{Li2017LearningWF}
Zhizhong Li and Derek Hoiem.
\newblock Learning without forgetting.
\newblock In \emph{IEEE Trans. PAMI}, 2017.

\bibitem[Liu et~al.(2018)Liu, Masana, Herranz, Van~de Weijer, Lopez, and
  Bagdanov]{Liu2018RotateYN}
Xialei Liu, Marc Masana, Luis Herranz, Joost Van~de Weijer, Antonio~M Lopez,
  and Andrew~D Bagdanov.
\newblock Rotate your networks: Better weight consolidation and less
  catastrophic forgetting.
\newblock In \emph{ICPR}, 2018.

\bibitem[Liu et~al.(2020)Liu, Su, Liu, Schiele, and Sun]{mnemonics}
Yaoyao Liu, Yuting Su, An-An Liu, Bernt Schiele, and Qianru Sun.
\newblock Mnemonics training: Multi-class incremental learning without
  forgetting.
\newblock In \emph{CVPR}, pp.\  12245--12254. Computer Vision Foundation /
  {IEEE}, 2020.

\bibitem[Lopez-Paz \& Ranzato(2017)Lopez-Paz and
  Ranzato]{LopezPaz2017GradientEM}
David Lopez-Paz and Marc'Aurelio Ranzato.
\newblock Gradient episodic memory for continual learning.
\newblock In \emph{NIPS}, volume~30, pp.\  6467--6476, 2017.

\bibitem[Mallya et~al.(2018)Mallya, Davis, and Lazebnik]{Mallya2018PiggybackAA}
Arun Mallya, Dillon Davis, and Svetlana Lazebnik.
\newblock Piggyback: Adapting a single network to multiple tasks by learning to
  mask weights.
\newblock In \emph{ECCV}, 2018.

\bibitem[McCloskey \& Neal(1989)McCloskey and Neal]{mccloskeyC89}
M.~McCloskey and Neal.
\newblock Catastrophic interference in connectionist networks: The sequential
  learning problem.
\newblock In \emph{Psychology of Learning and Motivation}, volume~24, pp.\
  109--165, 1989.

\bibitem[Oh et~al.(2012)Oh, Choi, Lee, and Noh]{cache-less}
Yongseok Oh, Jongmoo Choi, Donghee Lee, and Sam~H Noh.
\newblock Caching less for better performance: balancing cache size and update
  cost of flash memory cache in hybrid storage systems.
\newblock In \emph{USENIX FAST}, volume~12, pp.\ ~25, 2012.

\bibitem[Parisi et~al.(2018)Parisi, Kemker, L.~Part, Kanan, and
  Wermter]{parisiKPK18}
German Parisi, Ronald Kemker, Jose L.~Part, Christopher Kanan, and Stefan
  Wermter.
\newblock Continual lifelong learning with neural networks: A review.
\newblock In \emph{Neural Networks}, 2018.

\bibitem[Peng et~al.(2020)Peng, Shi, Dai, Jin, Ma, Xiong, Yang, and
  Qian]{Capuchin_asplos20}
Xuan Peng, Xuanhua Shi, Hulin Dai, Hai Jin, Weiliang Ma, Qian Xiong, Fan Yang,
  and Xuehai Qian.
\newblock Capuchin: Tensor-based gpu memory management for deep learning.
\newblock In \emph{ASPLOS}, pp.\  891--905. ACM, 2020.

\bibitem[Prabhu et~al.(2020)Prabhu, Torr, and Dokania]{gdumb}
Ameya Prabhu, Philip~HS Torr, and Puneet~K Dokania.
\newblock {GDumb}: A simple approach that questions our progress in continual
  learning.
\newblock In \emph{ECCV}, 2020.

\bibitem[Ratcliff(1990)]{er}
Roger Ratcliff.
\newblock Connectionist models of recognition memory: Constraints imposed by
  learning and forgetting functions.
\newblock In \emph{Psychological Review}, volume~97, pp.\  285--308, 1990.

\bibitem[Rebuffi et~al.(2017)Rebuffi, Kolesnikov, Sperl, and Lampert]{icarl}
Sylvestre-Alvise Rebuffi, Alexander Kolesnikov, Georg Sperl, and Christoph~H.
  Lampert.
\newblock {iCaRL}: Incremental classifier and representation learning.
\newblock In \emph{CVPR}, 2017.

\bibitem[Ren et~al.(2021)Ren, Luo, Wu, Zhang, Jeon, and Li]{sentinel}
Jie Ren, Jiaolin Luo, Kai Wu, Minjia Zhang, Hyeran Jeon, and Dong Li.
\newblock Sentinel: Efficient tensor migration and allocation on heterogeneous
  memory systems for deep learning.
\newblock In \emph{IEEE HPCA}, pp.\  598--611. {IEEE}, 2021.

\bibitem[Rhu et~al.(2016)Rhu, Gimelshein, Clemons, Zulfiqar, and
  Keckler]{vDNN_micro16}
Minsoo Rhu, Natalia Gimelshein, Jason Clemons, Arslan Zulfiqar, and Stephen~W.
  Keckler.
\newblock vdnn: Virtualized deep neural networks for scalable, memory-efficient
  neural network design.
\newblock In \emph{MICRO}, pp.\  1--13. {IEEE} Computer Society, 2016.

\bibitem[Seff et~al.(2017)Seff, Beatson, Suo, and Liu]{Seff2017ContinualLI}
Ari Seff, Alex Beatson, Daniel Suo, and Han Liu.
\newblock Continual learning in generative adversarial nets.
\newblock \emph{arXiv preprint arxiv:1705.08395}, 2017.

\bibitem[Shim et~al.(2021)Shim, Mai, Jeong, Sanner, Kim, and Jang]{aser}
Dongsub Shim, Zheda Mai, Jihwan Jeong, Scott Sanner, Hyunwoo Kim, and Jongseong
  Jang.
\newblock Online class-incremental continual learning with adversarial shapley
  value.
\newblock In \emph{AAAI}, volume~35, pp.\  9630--9638, 2021.

\bibitem[Shin et~al.(2017)Shin, Lee, Kim, and Kim]{Shin2017ContinualLW}
Hanul Shin, Jung~Kwon Lee, Jaehong Kim, and Jiwon Kim.
\newblock Continual learning with deep generative replay.
\newblock In \emph{NIPS}, 2017.

\bibitem[van~de Ven \& Tolias(2018)van~de Ven and Tolias]{ven2018three}
Gido~M van~de Ven and Andreas~S Tolias.
\newblock Three continual learning scenarios and a case for generative replay.
\newblock In \emph{NIPS Workshop on Continual Learning}, 2018.

\bibitem[Vitter(1985)]{vitter1985random}
Jeffrey~S Vitter.
\newblock Random sampling with a reservoir.
\newblock In \emph{ACM TOMS}, volume~11, pp.\  37--57, 1985.

\bibitem[Wang et~al.(2018)Wang, Ye, Zhao, Wu, Li, Song, Xu, and
  Kraska]{SuperNeurons_ppopp18}
Linnan Wang, Jinmian Ye, Yiyang Zhao, Wei Wu, Ang Li, Shuaiwen~Leon Song,
  Zenglin Xu, and Tim Kraska.
\newblock Superneurons: Dynamic gpu memory management for training deep neural
  networks.
\newblock In \emph{PPoPP}, pp.\  41--53, 2018.

\bibitem[Wang et~al.(2019)Wang, Jiang, Chen, Xu, Zhao, Lin, and Wang]{e2train}
Yue Wang, Ziyu Jiang, Xiaohan Chen, Pengfei Xu, Yang Zhao, Yingyan Lin, and
  Zhangyang Wang.
\newblock E2-train: Training state-of-the-art cnns with over 80\% energy
  savings.
\newblock In \emph{NIPS}, pp.\  5139--5151. Curran Associates, Inc., 2019.

\bibitem[Welling(2009)]{Welling2009herding}
Max Welling.
\newblock Herding dynamical weights to learn.
\newblock In \emph{ICML}, pp.\  1121--1128, 2009.

\bibitem[Wu et~al.(2018)Wu, Herranz, Liu, Wang, Van~de Weijer, and
  Raducanu]{Wu2018MemoryRG}
Chenshen Wu, Luis Herranz, Xialei Liu, Yaxing Wang, Joost Van~de Weijer, and
  Bogdan Raducanu.
\newblock {Memory Replay GANs}: learning to generate images from new categories
  without forgetting.
\newblock In \emph{NIPS}, 2018.

\bibitem[Wu et~al.(2019)Wu, Chen, Wang, Ye, Liu, Guo, and Fu]{bic}
Yue Wu, Yinpeng Chen, Lijuan Wang, Yuancheng Ye, Zicheng Liu, Yandong Guo, and
  Yun Fu.
\newblock Large scale incremental learning.
\newblock In \emph{CVPR}, 2019.

\bibitem[Zaharia et~al.(2010)Zaharia, Chowdhury, Franklin, Shenker, Stoica,
  et~al.]{spark}
Matei Zaharia, Mosharaf Chowdhury, Michael~J Franklin, Scott Shenker, Ion
  Stoica, et~al.
\newblock Spark: Cluster computing with working sets.
\newblock In \emph{USENIX Workshop on HotCloud}, pp.\ ~95, 2010.

\bibitem[Zenke et~al.(2017)Zenke, Poole, and Ganguli]{Zenke2017ContinualLT}
Friedemann Zenke, Ben Poole, and Surya Ganguli.
\newblock Continual learning through synaptic intelligence.
\newblock In \emph{ICML}, 2017.

\end{thebibliography}
}
\newpage 
\appendix
\section{Appendix}


\subsection{Implementation Details}
\label{sec:Impledetails}

First, we describe implementation details about the two components of the proposed method: swap
worker and episodic memory. 
Then, we describe the details about PyTorch integration of our implementation for ease of use.

\Paragraph{Swap worker.} \ours implements the swap worker through
multiprocessing~\citep{pythonmulti} in popular Python standard library so that
data swapping is running in parallel with PyTorch's default fetch workers
dedicated to data decoding and augmentation. The swap worker uses
\emph{asyncio}~\citep{asyncio} to asynchronously load samples from storage to memory,
effectively overlapping high-latency I/O operations with other \ours-related operations,
such as image decoding, sample replacement on EM, and entropy calculation.
The swap worker issues multiple data swapping requests without spinning on or
being blocked by I/O. As a result, it is sufficient to have only one swap
worker for \ours in the system.

\Paragraph{Episodic memory.} There are various ways to implement EM to be
shared between fetch workers and the swap worker. The current system favors
flexibility over performance, so we opt for implementing EM as a shared
object provided by \emph{Manager}~\citep{manager} in the Python standard library
(\emph{multiprocessing.managers}), which is based on
message passing in the server-client semantics. In terms of
flexibility, the Manager does not require the clients (\ie, fetch workers and
swap workers) to define the exact data layout in the EM address space or
coordinate for potential memory resizing to accommodate raw samples of
different sizes (\eg, image resolutions). Hence, it is sufficient for the
client workers to perform reads and writes on EM using indexes on the EM samples. An
alternative yet obviously higher-performance implementation would be using
\emph{multiprocessing.shared\_memory}~\citep{sharedmem}, which enables direct
reads and writes on EM by exposing a common region of memory to the
processes. Despite good performance, this method is less flexible as all
processes should be aware of the data layout in a designated EM address range
precisely at runtime, thus requiring additional coordination for sample
lookups and EM resizing. As our system evolves, we ultimately want to
combine the best of both methods to promise both flexibility and
performance.

\subsection{Datasets}
\label{sec:Datasets}

Each baseline is evaluated on its
own dataset used in the original work. The first rows of
Table~\ref{datasetcifar} and Table~\ref{datasetimagenet} show datasets used
in the CIFAR subset and the ImageNet subset, respectively, for all baselines.
ImageNet-100 is a ImageNet ILSVRC2012 subset used in iCaRL, which contains
images in the same resolution as those in the original ImageNet ILSVRC2012.
Other datasets used as the ImageNet subset have smaller image resolution than
the original one (\eg, 64$\times$64 for Tiny-ImageNet, 84$\times$84 for
Mini-ImageNet). In addition, we trained all baselines on ImageNet-1000 to verify the effectiveness of \ours on a large-scale dataset. We note that only ER, iCaRL, and BiC have been compared using the ImageNet-1000 dataset in the literature~\citep{bic}.

Datasets are split as done in the original work.
The second and third rows of Table~\ref{datasetcifar} and
Table~\ref{datasetimagenet} show the detailed information on the splitting
strategy. For all baselines, the ImageNet-1000 dataset is split into 10 tasks, each with 100 classes.
Note that all datasets are non-blurry, meaning that each task
consists of its own set of classes and samples belonging to a previous task
never appear in the next tasks. Since the experimental results are highly
sensitive to the class order in the continuous tasks to train, we follow the
same class order used in the original works.

\subsection{Experimental Details}
\label{sec:Expdetails}

We present the
effectiveness of the proposed \ours using seven CL methods of their own
setups. This section discusses detailed settings for each method so that the
results are reproducible by our source code. We first describe the metrics
used for the evaluations.

\begin{table}[t]
\centering
\resizebox{0.99\textwidth}{!}{
\caption{Dataset formation of CIFAR subset.}
\label{datasetcifar}
\begin{tabular}{lccccccc}
\toprule
& ER & iCaRL & TinyER & BiC & GDumb & DER++ & RM\\
\midrule
Dataset & CIFAR-100 & CIFAR-100 & CIFAR-100 & CIFAR-100 & CIFAR-10 & CIFAR-10 & CIFAR-10 \\
\# of Tasks & 10 & 10 & 20 & 10 & 5 & 5 & 5\\
\# of Classes per Task & 10 & 10 & 5 & 10 & 2 & 2 & 2\\
\bottomrule
\end{tabular}
}
\end{table}

\begin{table}[!t]
\centering
\resizebox{0.99\textwidth}{!}{
\caption{Dataset formation of ImageNet subset.}
\label{datasetimagenet}
\begin{tabular}{lccccccc}
\toprule
& ER & iCaRL & TinyER & BiC & GDumb & DER++ & RM\\
\midrule
Dataset &  ImageNet-100 & ImageNet-100 & Mini-ImageNet & ImageNet-100 & Tiny-ImageNet & Tiny-ImageNet & Tiny-ImageNet \\
\# of Tasks & 10 & 10 & 20 & 10 & 10 & 10 & 10\\
\# of Classes per Task & 10 & 10 & 5 & 10 & 20 & 20 & 20\\
\bottomrule
\end{tabular}
}
\end{table}

\subsubsection{Metrics}
\label{sec:Metrics}

\Paragraph{Final accuracy.} Final accuracy is an average accuracy over all
classes observed after the last task training is done.

\Paragraph{Final forgetting.} Forgetting indicates how much each
task has been forgotten while training new tasks~\citep{rwalk}. Forgetting for a task is
calculated by comparing the best accuracy observed over task insertions to
the final accuracy of the task when training is over. Final forgetting is an
average forgetting across all tasks when training is over.

\subsubsection{Baseline Details}
\label{sec:baselines}

\begin{itemize}
    \item \textbf{ER}~\citep{er} combines all samples in the current stream
        buffer and the current EM, and passes them over to the model as a
        training set, \ie, training bundle. There is no algorithmic
        optimization applied to the model itself. We manage EM as a ring
        buffer that assigns EM space equally over all classes observed so
        far. We use the same hyper-parameters and loss function (binary
        cross-entropy loss) as used in iCaRL.
    \item \textbf{iCaRL}~\citep{icarl} uses three algorithmic optimizations:
        distillation loss, herding, and nearest-mean-of-exemplar
        classification. To transfer the information of old tasks, iCaRL
        leverages the distillation loss using logits obtained from the most
        recently trained model for old classes: this loss information is
        considered as the ground truth for old classes. Herding is its own
        EM management method, which populates the samples whose feature
        vectors are the closest to the average feature vector overall
        stream data for each class. iCaRL allocates EM space equally overall observed classes.
    \item \textbf{TinyER}~\citep{tiny} explores four EM management
        strategies named reservoir, ring buffer, k-means, and mean of
        features. We adopt the reservoir in the experiments because it
        shows overall the highest performance in the original paper.
        Similar to ER, TinyER retrieves old samples from EM without other
        optimizations on the model itself. TinyER is batch-level learning
        and focuses on an extremely online setup that takes a single pass
        for every streamed batch.
    \item \textbf{BiC}~\citep{bic} runs bias correction on the last layer of the
        neural network, structured as fully connected layer, to mitigate
        data imbalance problem between old samples and new samples. The
        data imbalance is an inherent problem due to the limited size of EM,
        and it gets worse as we have a larger number of consecutive classes
        to train. Similar to iCaRL, BiC opts for distillation loss, but its
        entire loss function is a mixture of distillation loss and
        cross-entropy loss that is directly calculated from some reserved
        samples for old classes.
    \item \textbf{GDumb}~\citep{gdumb} is a simple rehearsal-based method
        that uses only the memory to train the model. The memory management
        is done via greedy balanced sampling, where GDumb tries to keep
        each class balanced by evicting data categorized into the majority
        class out of EM. Unlike other methods, the model is trained from
        scratch for inference and then discarded every time the memory is
        updated. GDumb uses cosine annealing learning-rate scheduler and
        cross-entropy loss for gradient descend.
    \item \textbf{DER++}~\citep{darker} is one of rehearsal-based methods
        with knowledge distillation. Unlike other methods, this approach
        retains logits (along with samples) in EM for knowledge
        distillation. For knowledge distillation, DER++ calculates
        euclidean distance between the logits stored in EM and the logits
        generated by the current network. To enable data swapping on DER++,
        we store the logits in the storage along with samples.
    \item \textbf{RM}~\citep{rainbow} uses the same backbone as GDumb, but
        it improves memory update policy and training method over GDumb.
        For memory management, RM calculates the uncertainty of each sample
        and tries to fill the memory with samples in a wide spectrum that
        ranges from robust samples with low uncertainity to fragile samples with high uncertainity  while keeping
        the classes balanced. In addition, data augmentation (DA) is
        proposed to advance the original RM implementation. We use RM
        without DA to apply data swapping in our work, but we include some
        results of RM with DA in Section~\ref{sec:RMwithDA}.
\end{itemize}

\Paragraph{Reproduction.} We use reported numbers from the original paper
for DER++ on Tiny-ImageNet~\citep{darker}. For iCaRL, we believe we
faithfully implement its details, but could not reach the accuracy reported
in the paper. 

As far as we know, there is no PyTorch source code that
reproduces iCaRL on both CIFAR-100 and ImageNet-100 datasets. In our
implementation for iCaRL, we refer to a PyTorch version written by the PodNet
authors~\citep{podnet} as they achieve the most comparable results. We use the
results obtained from the referred version rather than the reported results, because compared to
the reported accuracy, the obtained accuracy is nearly the same for CIFAR-100 and higher for ImageNet-100.

\begin{table}[ht]
\centering
\resizebox{0.99\textwidth}{!}{
\caption{Hyper-parameters of Table 1 in the main paper for CIFAR subset.}
\label{paramcifar}
\begin{tabular}{lccccccc}
\toprule
 & ER & iCaRL & TinyER & BiC & GDumb & DER++ & RM\\
\midrule
Learning Level & Task & Task & Batch & Task & Task & Task & Task\\
Model & ResNet32 & ResNet32 & ReducedResNet18 & ResNet32 & ResNet18 & ResNet18 & ResNet18\\
\# Passes per Bundle & 70 & 70 & 1 & 250 & 256 & 50 & 256\\
EM Size & 2000 & 2000 & 300 & 2000 & 500 & 500 & 500 \\
Batch Size & 128 & 128 & 10 & 128 & 16 & 32 & 16\\
Learning Rate & 2.0 & 2.0 & 0.1 & 0.1 & 0.05 & 0.03 & 0.05\\
Weight Decay & 1e-5 & 1e-5 & 0 & 2e-4 & 1e-6 & 0 & 1e-6\\
TI / CI & CI & CI & TI & CI & CI & CI & CI\\
\bottomrule
\end{tabular}
}
\end{table}

\begin{table}[ht]
\centering
\resizebox{0.99\textwidth}{!}{
\caption{Hyper-parameters of Table 1 in the main paper for ImageNet subset.}
\label{paramimagenet}
\begin{tabular}{lccccccc}
\toprule
& ER & iCaRL & TinyER & BiC & GDumb & DER++ & RM\\
\midrule
Learning Level & Task & Task & Batch & Task & Task & Task & Task\\
Model & ResNet18 & ResNet18 & ResNet18 & ResNet18 & DenseNet100 & ResNet18 & DenseNet100\\
\# Passes per Bundle & 60 & 60 & 1 & 100 & 128 & 100 & 128\\
EM Size & 2000 & 2000 & 500 & 2000 & 4500 & 500 & 4500 \\
Batch Size & 128 & 128 & 10 & 256 & 16 & 32 & 16\\
Learning Rate & 2.0 & 2.0 & 0.1 & 0.1 & 0.05 & 0.03 & 0.05\\
Weight Decay &  1e-5 & 1e-5 & 0 & 1e-4 & 1e-6 & 0 & 1e-6\\
TI / CI & CI & CI & TI & CI & TI & CI & CI\\
\bottomrule
\end{tabular}
}
\end{table}

\begin{table}[ht]
\centering
\resizebox{0.99\textwidth}{!}{
\caption{Hyper-parameters of Table 1 in the main paper for ImageNet-1000.}
\label{paramimagenet1000}
\begin{tabular}{lccccccc}
\toprule
& ER & iCaRL & TinyER & BiC & GDumb & DER++ & RM\\
\midrule
Learning Level & Task & Task & Batch & Task & Task & Task & Task\\
Model & ResNet18 & ResNet18 & ResNet18 & ResNet18 & DenseNet100 & ResNet18 & DenseNet100\\
\# Passes per Bundle & 60 & 140 & 1 & 100 & 128 & 100 & 128\\
EM Size & 20000 & 20000 & 20000 & 20000 & 20000 & 20000 & 20000 \\
Batch Size & 128 & 128 & 32 & 256 & 16 & 16 & 16\\
Learning Rate & 2.0 & 2.0 & 0.1 & 0.1 & 0.05 & 0.03 & 0.05\\
Weight Decay &  1e-5 & 1e-5 & 0 & 1e-4 & 1e-6 & 0 & 1e-6\\
TI / CI & CI & CI & TI & CI & TI & CI & CI\\
\bottomrule
\end{tabular}
}
\end{table}

\subsubsection{Hyper-parameters}
\label{sec:Hyperparameters}

We follow hyper-parameters presented in the original works: we did not perform
hyper-parameter search for the baselines. Table~\ref{paramcifar},
Table~\ref{paramimagenet}, and Table~\ref{paramimagenet1000} present all the details on the hyper-parameters.

Although DER++ updates EM in batch-level and does not consider task boundary, for a larger dataset than MNIST, the original paper chooses to takes multiple passes per bundle. So, we deem DER++ to be a task-level learning method as long as we use CIFAR-100 and Tiny ImageNet as its training dataset. Here, TI and CI denote task-incremental learning and class-incremental learning, respectively. TI is an easy and simplified scenario, where the task ID is given at both training and inference. In TI setting, the model can classify the input among the classes that belong to the provided task ID. On the contrary, CI is the setting where the task ID is unknown during inference, which is a more realistic case than TI.

\subsubsection{Detailed Specification of Machines}
\label{sec:Machines}

Our experiments are performed on machines with HW specification 
as presented in Table~\ref{machinespec}. These machines are also used in measuring
the impact on training speed with data swapping.

\begin{table}[ht]
\centering
\caption{Machine Specfication.}
\label{machinespec}
\begin{tabular}{lc}
\toprule
& Machine Specs\\
\midrule
CPU & Intel(R) Xeon(R) Gold 6226 CPU @ 2.70~GHz $\times$ 2 \\

GPU & NVIDIA Geforce RTX 2080Ti (11~GB) $\times$ 4 \\

RAM & 128~GB, 2666~MHz \\

SSD  & Intel SSD D3 Series 480~GB \\

HDD  & Western Digital Ultrastar DC HC310 4~TB \\

\bottomrule
\end{tabular}
\end{table}

\subsection{Additional Results}
\label{sec:Addres}

\subsubsection{Distillation Analysis}
\label{sec:distill_detail}

\begin{figure}[ht]
\center
\includegraphics[width=0.5\linewidth]{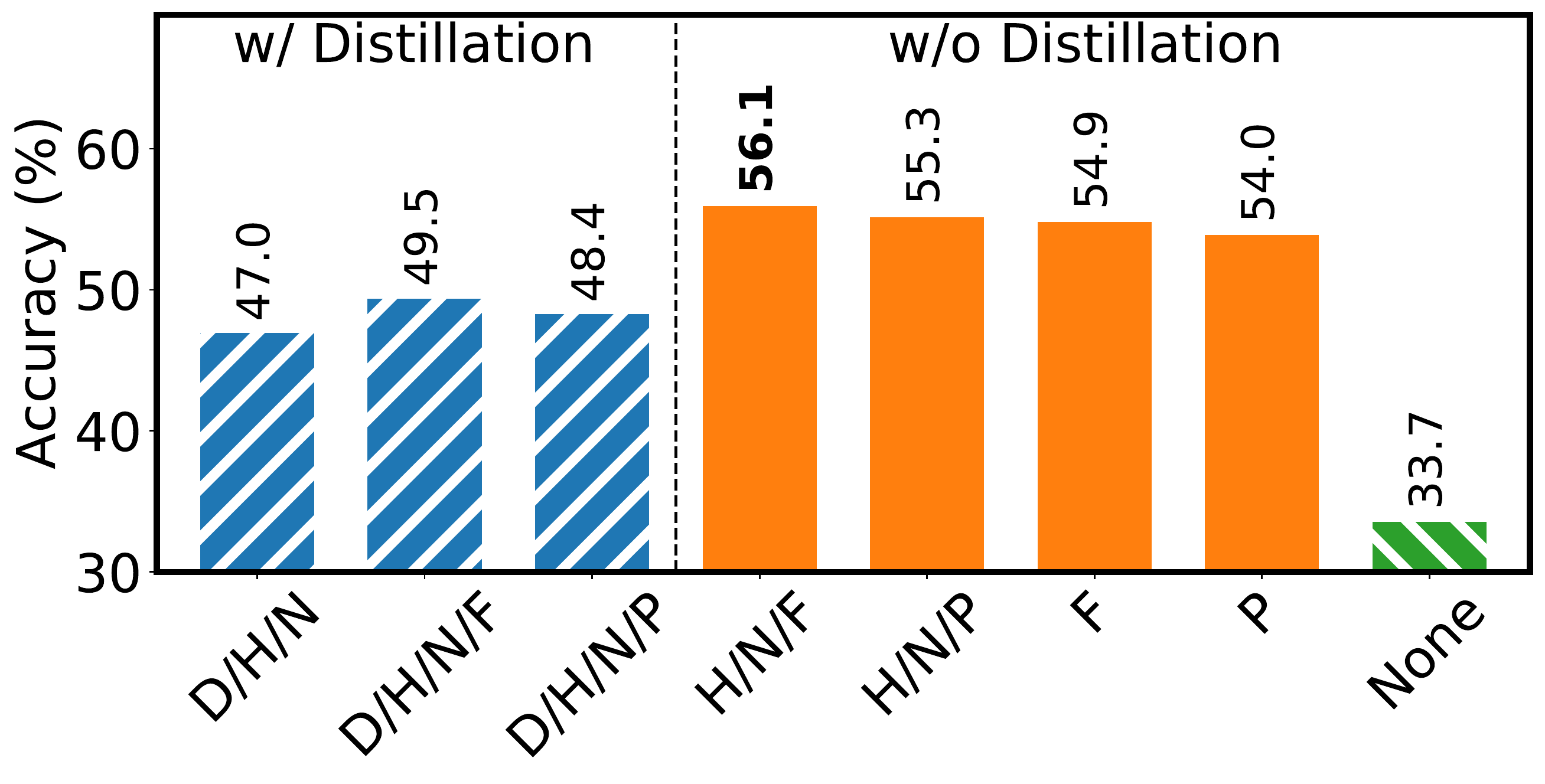}
\caption{Accuracy of eight combinations driven by algorithmic features in iCaRL and \ours.}
\label{fig:icarl_distill}
\end{figure}

\Paragraph{Effectiveness of features of iCaRL on \ours.} We explore iCaRL by measuring
accuracy for all possible 32 combinations based on its algorithmic features,
\ie, knowledge distillation (\textbf{D}), herding (\textbf{H}), and
nearest-mean-of-exemplars (\textbf{N}), along with our \ours-100 (\textbf{F})
or \ours-50 (\textbf{P}). In Figure~\ref{fig:icarl_distill}, we show eight
combinations that are sufficient to support the three interesting findings.
First, data swapping without distillation (orange bars) outperforms the other
combinations including pure iCaRL (blue and green bars). Second, for
combinations with distillation, applying data swapping does not deliver great
accuracy (D/H/N \emph{vs} the other two in blue bars). Finally, data swapping
does not seem to necessitate sophisticated algorithmic features (F\&H
\emph{vs} D/H/N), inferring a model simplification potential for
episodic memory.

\begin{figure}[ht]
\center
\includegraphics[width=\linewidth]{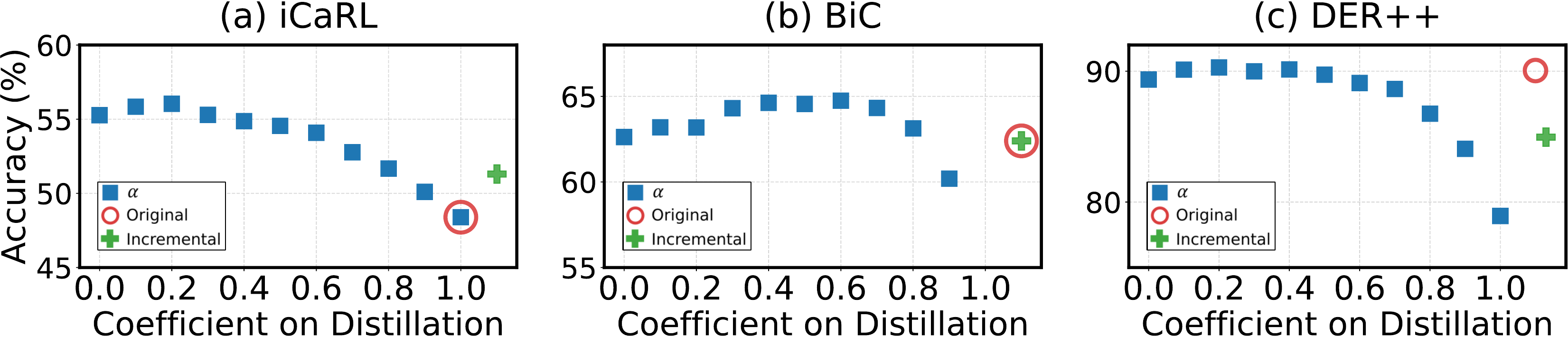}
\caption{
Accuracy of distillation-based methods with \ours-50 on CIFAR subset while varying coefficient ($\alpha$) values on the distillation loss in calculating training loss.
}
\label{fig:distill_cifar}
\end{figure}

\Paragraph{Knowledge distillation on \ours.}
Figure~\ref{fig:distill_cifar} show accuracy for CIFAR subset while varying $\alpha$ values in $\alpha \times soft\ label\ loss + (1-\alpha) \times hard\ label\
loss$ for iCaRL, BiC, and DER++. We can make the same conclusions as discussed in the `Knowledge distillation on \ours' paragraph of Section~\ref{sec:Results}. Below, we describe how each distillation-based method can be transformed into the presented model for loss calculation.

The original loss function of iCaRL~\citep{icarl} is defined as:

\begin{equation}
\begin{split}
\mathcal{L}_{icarl}(x_i) = -[\sum_{y=s}^{t}\{\delta_{y = y_i}\log g_y(x_i)+\delta_{y \neq y_i}\log(1-g_y(x_i))\}\\
+\sum_{y=1}^{s-1}\{q_i^y\log g_y(x_i)+(1-q_i^y)\log(1-g_y(x_i))\}]
\end{split}
\end{equation}
\label{eq:icarl}

where ${q_i^y}$ is the output of the old model, ${g_y(x_i)}$ is the output of the current model, $\{1,2,..,s-1\}$ is a set of old classes and $\{s,...,t\}$ is the set of new classes.

For distillation, it uses soft targets from the previous model for old classes of all current training set. As a result, training the current model heavily relies on the performance of the previous model. Especially, when data that belongs to old classes replays, since the target of loss is only soft output from the previous model, it is likely that the similar soft output from the old model is repeatedly distilled without the correct hard label. Due to such aggressive distillation, iCaRL cannot take an advantage of CarM, which enables to replay and train abundant old data, hindering positive decision boundary corrections. That is, the wrongly predicted samples from the old model will be predicted wrongly also in the future even if they are replayed several times by CarM. BiC and DER++ use distillation loss, however, unlike iCaRL, they provide a loss term of which target for old classes is ground truth, the correct hard label. As a result, BiC and DER++ could get higher accuracies with CarM. To evaluate Figure~\ref{fig:distill} and Figure~\ref{fig:distill_cifar}, we modified the loss function of iCaRL, adding another binary cross entropy that uses the ground truth as the target, which is referred to hard label loss, as following:

\begin{equation}
\begin{split}
\mathcal{L}_{modified}(x_i) = \alpha\mathcal{L}_{icarl}(x_i)-(1-\alpha)\sum_{y=1}^{t}\{\delta_{y = y_i}\log g_y(x_i)+\delta_{y \neq y_i}\log(1-g_y(x_i))\}
\end{split}
\end{equation}
\label{eq:modified}

Since BiC and DER++ already have its own hard label loss, we did not modify loss function.
Note that when $\alpha$ is set to 1.0 in BiC, it is unable to train any new data, which is unrealistic situation. So we excluded the result of $\alpha=1.0$ of BiC.

\begin{table}[ht]
\centering
\scriptsize
\resizebox{0.99\textwidth}{!}{
\begin{tabular}{lccccccc}
    \toprule
    Method & ER & iCaRL & TinyER & BiC & GDumb & DER++ & RM\\
    \midrule
    \textbf{Random}   & 54.08$\pm$0.50 & 48.27$\pm$0.51 & 59.64$\pm$1.91 & 62.53$\pm$0.25 & 52.39$\pm$1.80 & 89.82$\pm$0.11 & 66.91$\pm$0.81\\
    \textbf{Entropy}   & 54.00$\pm$0.49 & 48.39$\pm$0.41 & 59.99$\pm$2.12 & 62.40$\pm$0.40 & 52.64$\pm$1.64 & 90.05$\pm$0.38 & 66.66$\pm$0.78\\
    \textbf{Dynamic}  & 54.16$\pm$0.37 & 48.39$\pm$0.57 & 59.62$\pm$0.93 & 62.32$\pm$0.24 & 52.92$\pm$2.09 & 89.92$\pm$0.22 & 66.61$\pm$0.73\\
    \bottomrule
\end{tabular}
}
\vspace{-1em}
\caption{Comparison of accuracy for data swapping policies for \ours-50 on CIFAR subset.
\vspace{-1em}}
\label{tab:policies_cifar}
\end{table}

\subsubsection{Incremental Accuracy of Table~1 in the Main Paper}
\label{sec:incacctable1}

\Paragraph{Incremental accuracy.} We here report
incremental accuracy as an additional performance metric. Incremental accuracy is a set of average accuracy over classes observed so far after training each task. 

\begin{figure*}[]
\center
\includegraphics[width=\linewidth]{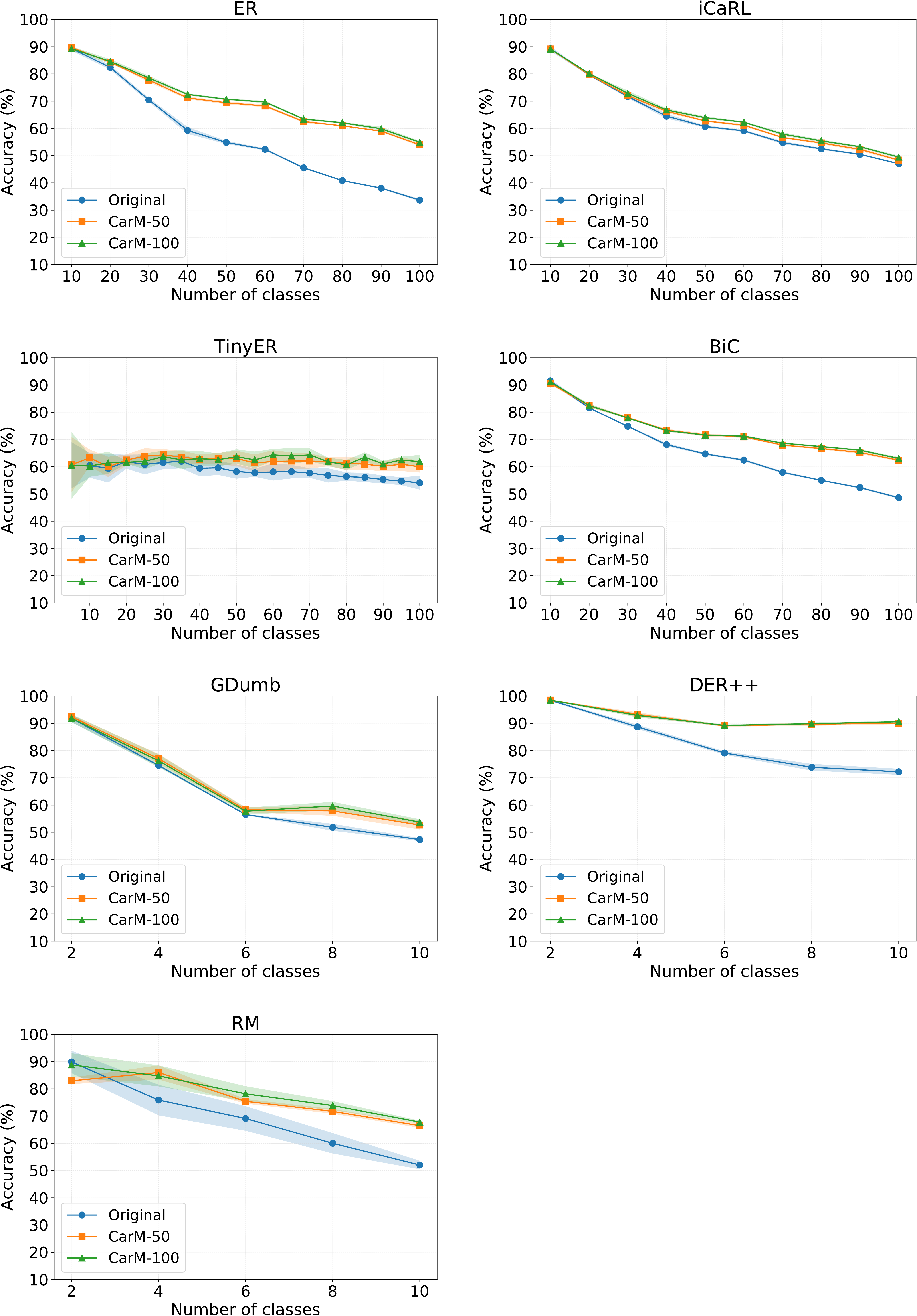}
\caption{Incremental accuracy on CIFAR subset.}
\label{fig:inc_acc_cifar}
\end{figure*}

Figure~\ref{fig:inc_acc_cifar} and Figure~\ref{fig:inc_acc_cifar} show the incremental accuracy of Table~1 in the main paper. We also mark the accuracy from original paper of iCaRL on CIFAR-100, iCaRL on ImageNet-100, BiC on ImageNet-100 and DER++ on Tiny-ImageNet. In general, the more tasks (classes) come, the larger gap of accuracy between original and \ours. This implies that running on \ours could better mitigate the catastrophic forgetting for long-term training.

\subsubsection{Ablation Study on ER, iCaRL, and GDumb}
\label{sec:ssicarlgdumbrm}

\begin{figure}[ht]
\center
\includegraphics[width=0.6\linewidth]{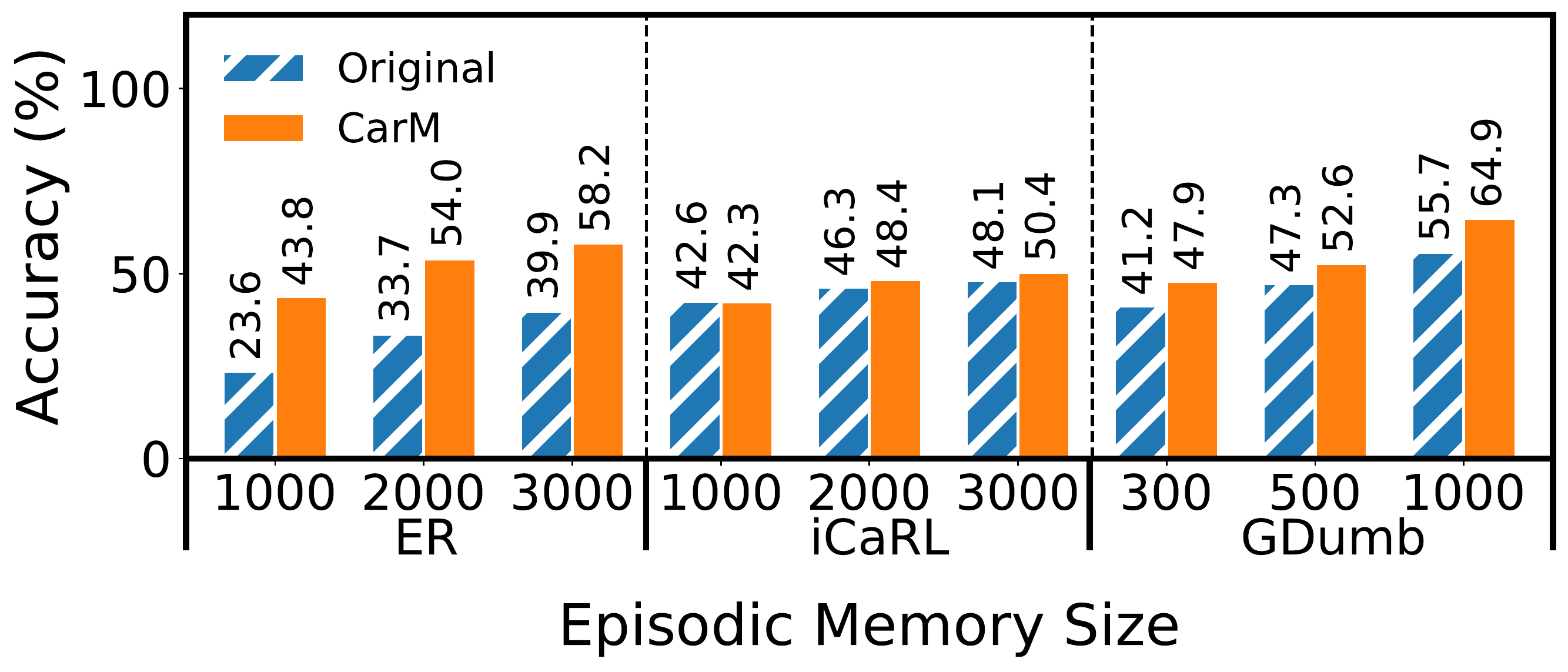}
\caption{Accuracy over varying EM sizes.}
\label{fig:buffer_size_supp}
\end{figure}

\begin{figure}[ht]
\centering %
\floatsetup{floatrowsep=qquad}
\begin{floatrow}[3]%
\ffigbox[\FBwidth]{\caption{Accuracy over varying data swapping ratios.}\label{fig:swap_ratio_supp}}{\includegraphics[width=0.3\textwidth]{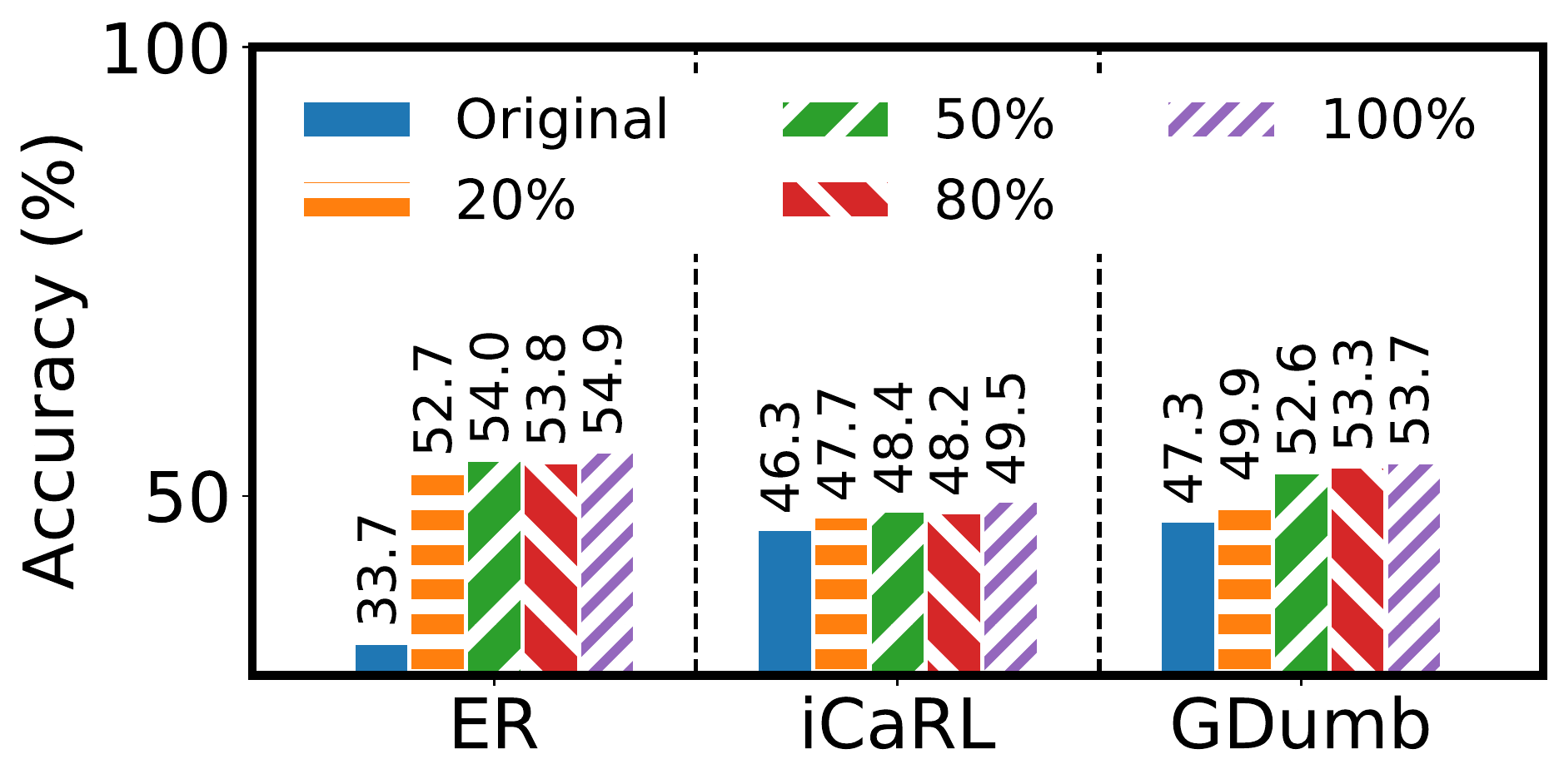} }
\ffigbox[\FBwidth]{\caption{Accuracy over varying storage capacities.}\label{fig:storage_supp}}{\includegraphics[width=0.3\textwidth]{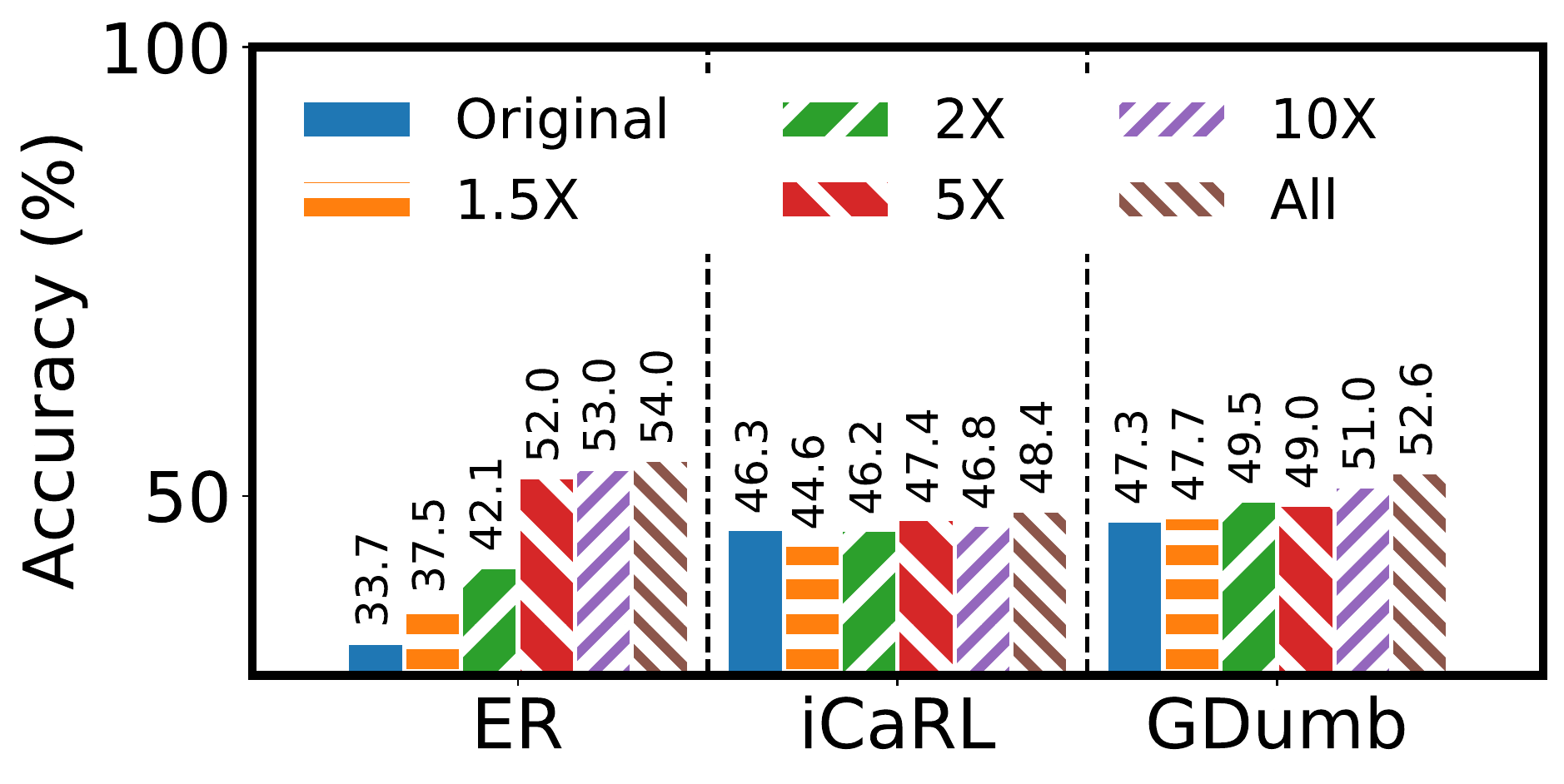}}
\ffigbox[\FBwidth]{\caption{Accuracy with learning 50 tasks.}\label{fig:swap_policy_supp}}{\includegraphics[width=0.3\textwidth]{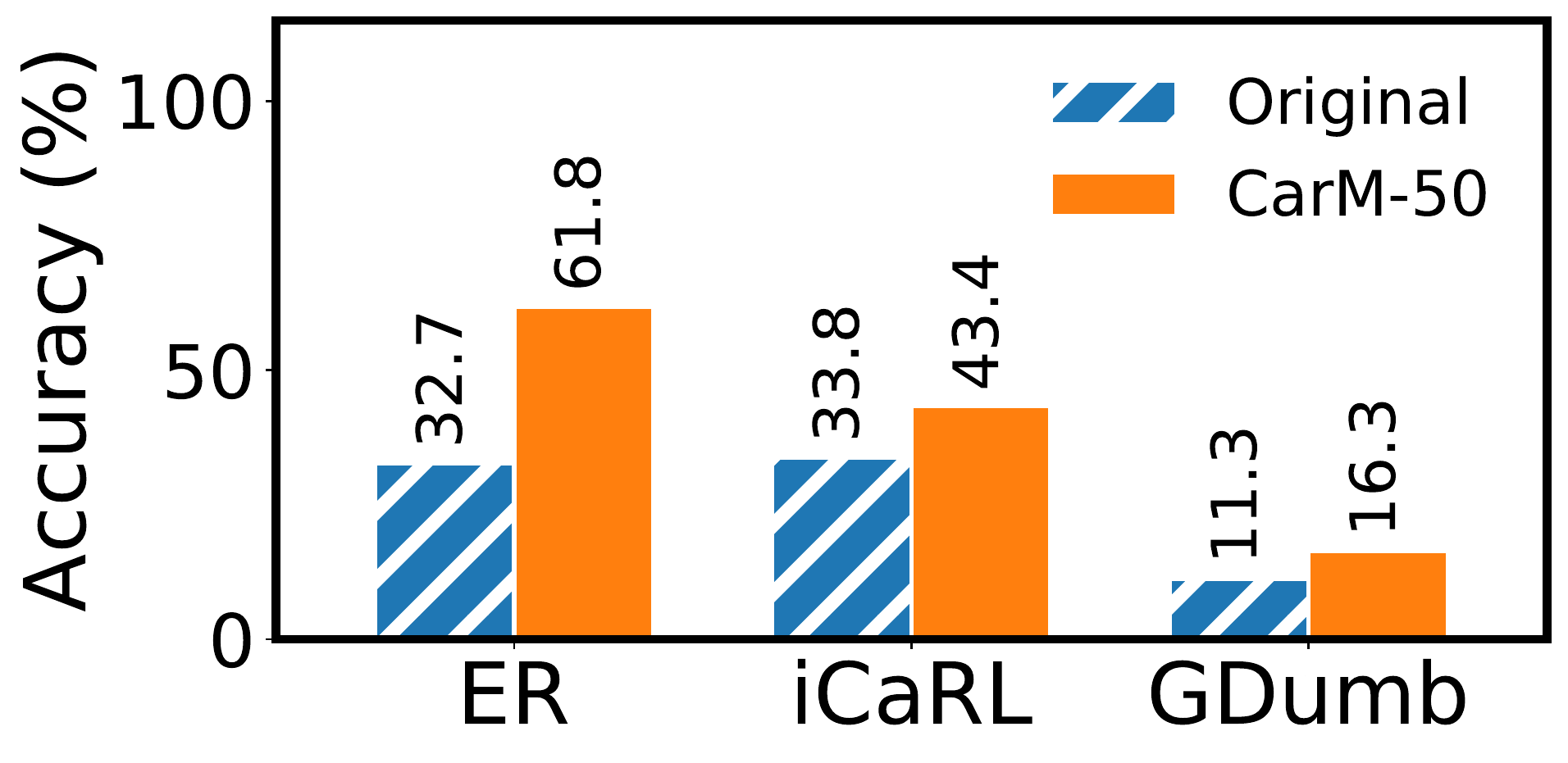}}
\end{floatrow}
\end{figure}

We report the results for an ablation study on ER, iCaRL, and GDumb, which were not presented in the main paper. Figure~\ref{fig:buffer_size_supp} shows accuracy over varying EM sizes, Figure~\ref{fig:swap_ratio_supp} shows accuracy over varying swapping ratios, Figure~\ref{fig:storage_supp} shows accuracy over varying storage capacity, and Figure~\ref{fig:swap_policy_supp} shows accuracy with learning 50 tasks.
In general, we found the similar observations as discussed in Section~\ref{sec:Ablation}.


\subsubsection{Results of RM with Data Augmentation}
\label{sec:RMwithDA}

We implement RM with data augmentation and show the results in Table~\ref{RMwithDA} using CIFAR-10 dataset. Both \ours and \ours-50 improves accuracy significantly over the baseline method.

\begin{table}[ht]
\centering
\caption{Accuracy improvement of using \ours w.r.t. RM with Data Augmentation on CIFAR-10.}
\label{RMwithDA}
\begin{tabular}{lccc}
\toprule
& Original & CarM-50 & CarM-100\\
\midrule
Final Accuracy & 68.07$\pm$1.57 & 84.07$\pm$0.83 & 84.85$\pm$0.42 \\
Final Forgetting & 15.09$\pm$1.72 & 5.12$\pm$0.89 & 4.36$\pm$0.95 \\
\bottomrule
\end{tabular}
\end{table}

\begin{table}[t]
\centering
\scriptsize
\resizebox{0.9\textwidth}{!}{
\begin{tabular}{lccccccc}
    \toprule
    Method & ER & iCaRL & TinyER & BiC & GDumb & DER++ & RM\\
    \midrule
    \textbf{Original}   & 33.87 & 46.54 & 55.18 & 49.44 & 47.41 & 71.79 & 51.87\\
    \textbf{CarM-50}   & 53.80 & 48.29 & 58.45 & 62.30 & 52.68 & 89.98 & 67.11\\
    \textbf{CarM-100}  & 55.38 & 49.11 & 59.79 & 63.21 & 52.91 & 90.28 & 67.24\\
    \bottomrule
\end{tabular}
}
\vspace{-1em}
\caption{CarM using a NVIDIA Jetson TX2\vspace{-1em}}
\label{tab:jetson}
\end{table}

\subsubsection{\ours on Embedded Device}
\label{sec:JetsonResults}

We evaluate \ours using a NVIDIA Jetson TX2 to show its efficacy when running on a representative embedded AI computing device. Table~\ref{tab:jetson} shows all baselines with \ours-50 and \ours-100 on CIFAR subset. We see accuracy improvements with \ours as similarly observed in the main paper.

\subsection{Discussions}
\label{sec:Discussions}


We have taken early steps towards leveraging both memory and storage to overcome the forgetting problem in CL while preserving the same training efficiency, which we find to be effective for the hardware we tested. However, as the characteristics between the memory and storage may vary significantly, the storage access latency may still become a significant bottleneck unless carefully exploited. Ideally, given the specs of a hardware configuration (\eg, computation, memory, and available I/O bandwidth), the swapping mechanism could decide an optimal policy to increase the memory capacity without adding additional latency. We leave this as an area of future work, which would make CarM more robust and resilient to variations in different hardware settings.

\begin{figure*}[]
\center
\includegraphics[width=\linewidth]{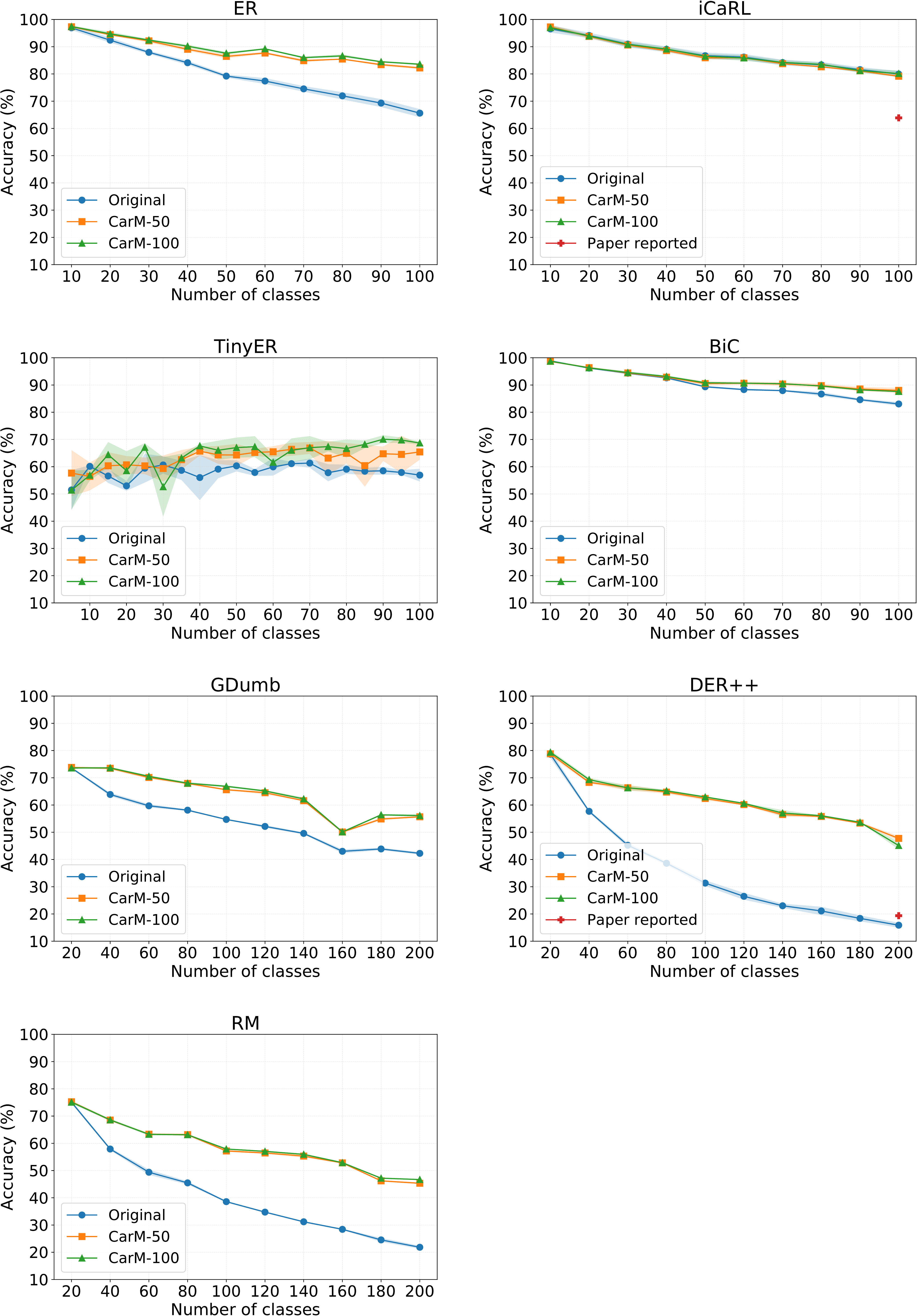}
\caption{Incremental accuracy on ImageNet subset.}
\label{fig:inc_acc_imagenet}
\end{figure*}

\end{document}